\RequirePackage{fix-cm}
\documentclass[smallextended]{svjour3}       
\smartqed  
\usepackage{graphicx}
\usepackage[ruled,vlined,linesnumbered]{algorithm2e}
\usepackage{ulem}
\usepackage{subfig}

\usepackage[dvipsnames]{color}

\usepackage[%
 centering,
 paperwidth=21cm, paperheight=29.7cm,
 textwidth=15cm, textheight=23cm,
 headheight=13.6pt,
 tmargin=80pt,
 hoffset=0mm, voffset=10mm,
 hmarginratio=1:1, vmarginratio=1:1,
 nomarginpar,
]{geometry}

\newcommand{\etal}{\textit{et al}.}
\newcommand{\ie}{\textit{i}.\textit{e}.}
\newcommand{\eg}{\textit{e}.\textit{g}.}

\def\x{\mathbf{x}}
\def\y{\mathbf{y}}

\def\u{\mathbf{u}}
\def\v{\mathbf{v}}

\begin{document}

\title{FALDOI: a new minimization strategy for large displacement variational optical flow \thanks{The first, second and third authors acknowledge partial support by MICINN project, reference  MTM2012-30772, by TIN2015-70410-C2-1-R (MINECO/FEDER, UE) and by GRC reference 2014 SGR 1301, Generalitat de Catalunya, and the fourth author by the European Research Council (advanced grant Twelve Labours), Office of Naval research (ONR grant N00014-14-1-0023).}
}


\author{Roberto P.Palomares         \and
        Enric Meinhardt-Llopis \and
        Coloma Ballester \and
        Gloria Haro
}


\institute{R.P.Palomares, C. Ballester and G. Haro \at
              Universitat Pompeu Fabra \\
              \email{roberto.palomares@upf.edu, coloma.ballester@upf.edu, gloria.haro@upf.edu}           
           \and
           E.Meinhardt-Llopis \at
               \'Ecole Normale Sup\'erieure de Cachan\\
              \email{enric.meinhardt@cmla.ens-cachan.fr}
}

\date{Received: date / Accepted: date}

\maketitle
\begin{abstract}
We propose a large displacement optical flow method that introduces a new strategy to compute a good local minimum of any optical flow energy functional. 
The method requires a given set of discrete matches, which can be extremely sparse, and an energy functional which locally guides the interpolation from those matches. In particular, the matches are used to guide a structured coordinate-descent of the energy functional around these keypoints. It results in a two-step minimization  method  at the finest scale which 
is very robust to the inevitable outliers of the sparse matcher and able to capture large displacements of small objects. 
Its benefits over other variational methods that also rely on a set of sparse matches are its robustness against very few matches, high levels of noise and outliers. We validate our proposal using several optical flow variational models. 
The results consistently outperform the coarse-to-fine approaches and achieve good  qualitative and quantitative performance on the standard optical flow benchmarks.
\end{abstract}
\section{Introduction}\label{sec:intro}

Optical flow is the apparent motion field between two consecutive frames of a video. More generally, it can be defined as 
 a dense correspondence field between an arbitrary pair of images.  There are two large families of methods for computing image correspondences: local and global methods. 
 Local methods establish a point correspondence by minimizing a distance measure between the matching neighborhoods \cite{gabriele_buades,kuk-jin}. They provide a sparse correspondence field since not all the image points are discriminative enough to guarantee a single correspondence. On the other hand, global or variational methods~\cite{Horn_Schunk:1981,alvarez2000reliable,Bro04a,Zach,Zimmeretal09,strekalovskiy-et-al-siims14} provide a dense solution by minimizing a global energy.
 Recent work on optical flow estimation~\cite{bailer2015flowICCV,chen2013large,fortun2014aggregation,kennedy2015optical,menze2015discrete,yang2015dense,EpicFlow,leordeanu2013locally} is mostly focused on solving the major challenges that appear in realistic scenarios and outdoor scenes, such as large displacements, motion discontinuities, illumination changes, and occlusions.
\begin{figure*}[htbp]
\centering
\subfloat[First frame and 4 initial seeds (in red)]{
    \includegraphics[width=0.4\textwidth]{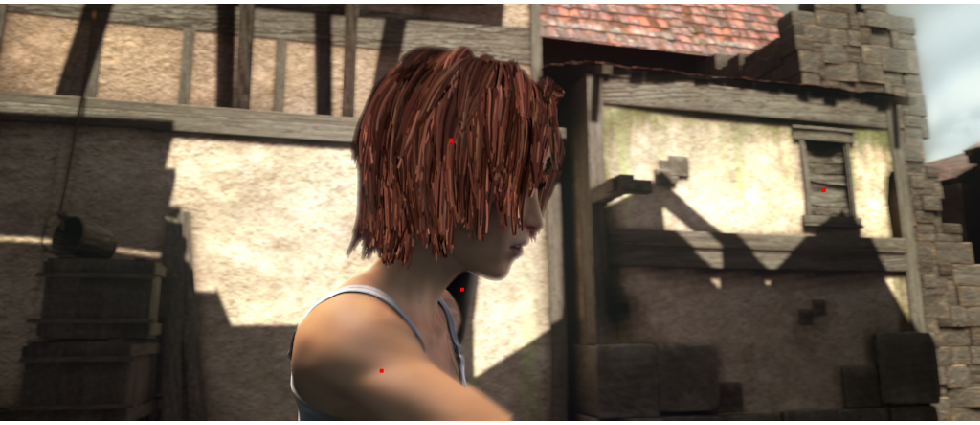}
    }
\subfloat[C. coding]{
    \includegraphics[width=0.1\textwidth]{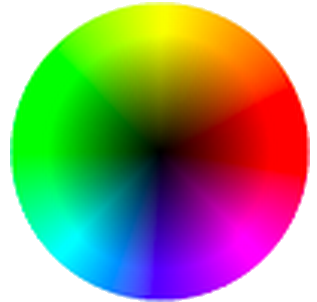}
  \label{fig:color_coding}}
  \subfloat[Second frame]{
    \includegraphics[width=0.4\textwidth]{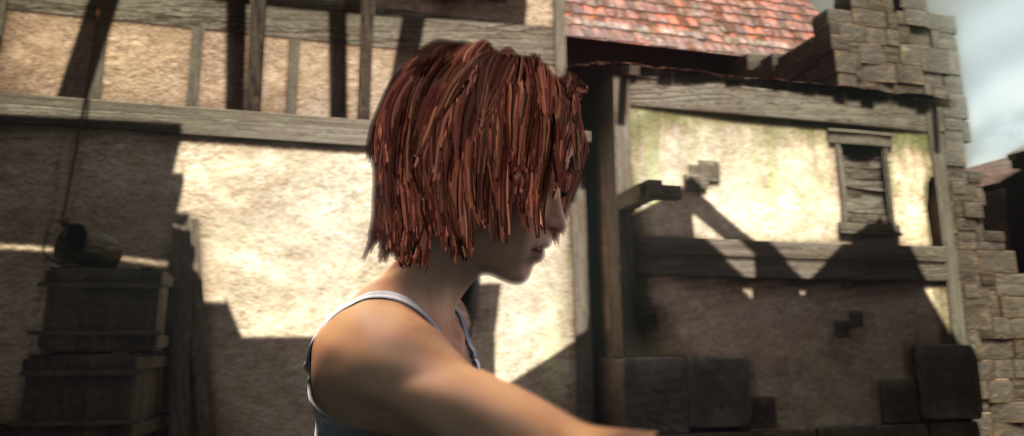}
  }
  \\
  \subfloat[Ground truth]{
    \includegraphics[width=0.45\textwidth]{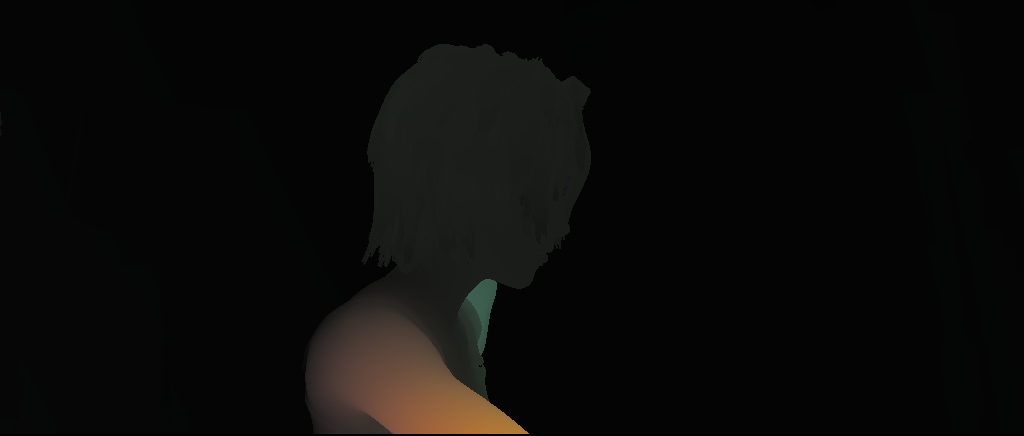}
  }
  \subfloat[Our result]{
    \includegraphics[width=0.45\textwidth]{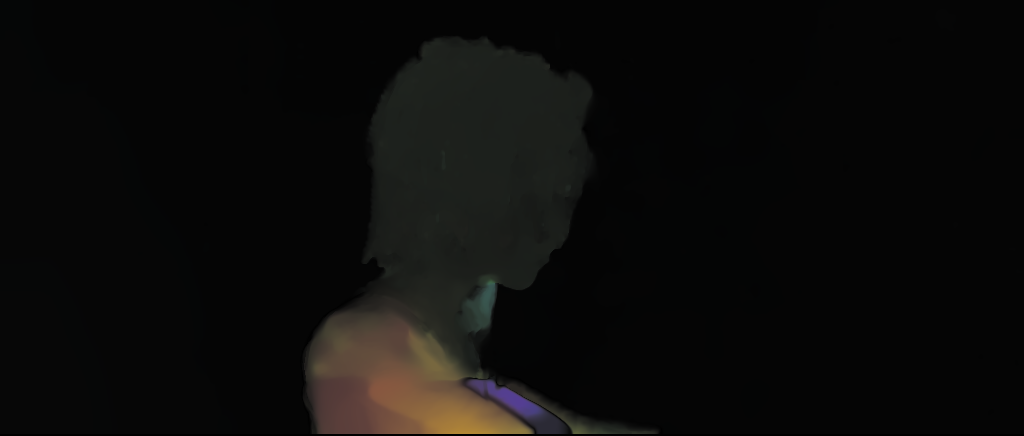}
  }
  \caption{Example illustrating that it is enough to have a single matching seed per each region with an expected smooth motion; 4 seeds in this case, which have been computed using SIFT. The first row shows the original pair of frames $I_t$ and $I_{t+1}$, where in $I_t$ we have superimposed in red the position of the 4 seeds (head, arm, dress and background). Last row shows our result and the ground truth. The energy functional used is the classical $TV_{\ell_2}$-$L1$.}
  \label{fig:sintel_claim_alley1}
\end{figure*}

 In this article we introduce a new strategy to compute good local minima of any
optical flow energy functional which allows to capture large displacements.  The method relies on a discrete set of matches
between the two input images, which 
can be extremely sparse and contaminated by outliers, and is used
as a guide to find a local minimum
of the chosen energy functional. 
The novelties include how these matches are used in the optimization problem.  
The proposed method is a two-step minimization process of the optical flow energy. 
The first step computes a good and dense local minimum by propagating the initial matches 
with a region growing strategy. The propagation is driven by the minimization of the energy on patches and the order of update is established by the value of the local version of the energy, \ie~the energy restricted to a 
square patch of the image domain.
 It can be thought as a grouped coordinate descent of the
target functional, where the variables are defined by the optical flow values in a patch, and the functional is minimized on the patch while fixing the rest of flow values. 
Coordinate descent methods provide efficient and fast optimization schemes even for huge-scale problems~\cite{Nesterov2012}. The basic idea is to minimize a multi-variable objective function by optimizing over a single coordinate while holding others fixed. 
The sweep pattern is the name used to define how the alternating minimization process moves along all the coordinates. In order to obtain a good minimum and accelerate convergence, we follow an adaptive choice of the sweep pattern driven by a seed growing algorithm based on the value of the energy after local minimization in a patch centered at that coordinate. This information is stored in a priority queue from which the coordinate is selected and the adaptive sweep pattern is iteratively created.
The result of the first step is a dense optical flow which is further refined in the second step: A global minimization of the energy taking as initial condition the flow obtained in the first stage. Both minimization steps work with the full resolution of the image directly.

Our method allows to choose the energy functional. For instance, by choosing it with desirable properties such as robustness to illumination changes, occlusions and motion discontinuities, those properties are inherited by our method. 
We can compare our method to the common coarse-to-fine (also called multi-scale) approach or to the strategy of including the information of sparse matches by incorporating an extra term in the energy that penalizes deviations from the flows given by the matches \cite{Brox_Malik_LD,Stoll2013,Weinzaepfel2013}; both of them also find local minima of arbitrary optical flow energy functionals. We find that we consistently outperform the multi-scale strategy 
for a variety of input  sparse matches and energy functionals. We present qualitative and quantitative results on several datasets such as Middlebury \cite{Middlebury}, MPI-Sintel \cite{Sintel}, KITTI 2012 \cite{Kitti}, and KITTI 2015 \cite{Menze2015CVPR} datasets. The performance is better than LDOF \cite{Brox_Malik_LD} and comparable to DeepFlow \cite{Weinzaepfel2013} while being more robust, less dependent on the density of seeds and based only  on a single energy with two terms (no need of extra parameters).
In our case, the ability to find large displacements requires only that {\sl at
least one} match is correctly given for each object in motion, as shown in the experiment of Figure~\ref{fig:sintel_claim_alley1}.  
Moreover, we find that our minimization strategy is very robust to the presence of many incorrect
keypoints in the input, as illustrated in Figure~\ref{fig:toy_outliers_sample}, where there are only 2 correct matches and 508 outliers. In contrast, the alternative strategy of including sparse matches in the energy functional with an additional term is not robust to a small number of matches (see Figure \ref{fig:hypermacaco}) nor to a large percentage of wrong matches or high levels of noise (see Table \ref{table:noise}). 

The remainder of the paper is organized as follows. In Section~\ref{sec:RelatedWork} we revise previous work on optical flow estimation. Section~\ref{sec:faldoi} presents our large displacement optical flow method (which we denote by FALDOI, as an acronym made of the first letter of the words
"large displacement optical flow method by an astute initialization"). 
Section~\ref{sec:results} presents results from several energy functionals 
and an analysis of the properties and performance of our strategy depending on the density of the initial set of discrete matches. Finally, the conclusions are summarized in Section~\ref{sec:Conclusions}. 

\section{Related work}
\label{sec:RelatedWork}
The seminal work of Horn and Schunk \cite{Horn_Schunk:1981} has inspired many subsequent works also based
on an energy-based formulation of the optical flow problem and focused on different kind
of limitations and their improvements. Initial progress was devoted to the use of more robust data terms and Total-Variation-based regularizers in order to obtain sharper solutions~\cite{Bro04a,black1996robust,sun2010secrets,Zach}.
The correct preservation of motion discontinuities is a key issue that has motivated many proposals in the smoothness term:
Starting from the use of decreasing functions of the image gradients~\cite{chen2013large,sanchez2014preserving,xu2012motion}, diffusion tensors~\cite{nagel1986investigation,werlberger2009anisotropic,zimmer2011optic},
 coupled regularization of the flow channels \cite{strekalovskiy-et-al-siims14,Palomares2014} 
 or second-order regularization \cite{ranftl2014non}, to the more recent non-local regularization terms \cite{krahenbuhl2012efficient,sun2014quantitative,werlberger2010motion}. 
The classical hypothesis for defining the data term has been the brightness constancy assumption but this is very limiting in realistic scenarios where illumination changes as well as occlusions may appear.  Robustness against additive illumination changes can be obtained by using the gradient constancy assumption \cite{papenberg2006highly}; while advanced data terms based on patch measures present more general invariances, in particular, the Normalized Cross Correlation \cite{Steinbruecker-et-al-vmv09,vogel2013evaluation,werlberger2010motion}  is invariant to linear brightness changes and the Census transform is invariant under monotonically increasing rescalings \cite{hafner2013census,muller2011illumination,stein2004efficient}. Patch-based data terms are used in the state-of-the-art methods, including stereo and optical flow problems. As they are more informative and allow to better characterise the local image structure they result in more accurate flow estimations. 
On the other hand, occlusions represent an important problem for optical flow estimation methods and some works include a characterization of the occlusion areas in the energy,~\cite{alvarez2002symmetrical,ayvaci2012sparse,BGLC,fortun2014aggregation}, while others  estimate the occlusion areas based on a triangulation of the image~\cite{kennedy2015optical}. 
\begin{figure*}[htbp]
\centering
  \subfloat[First frame]{
    \includegraphics[width=0.3\textwidth]{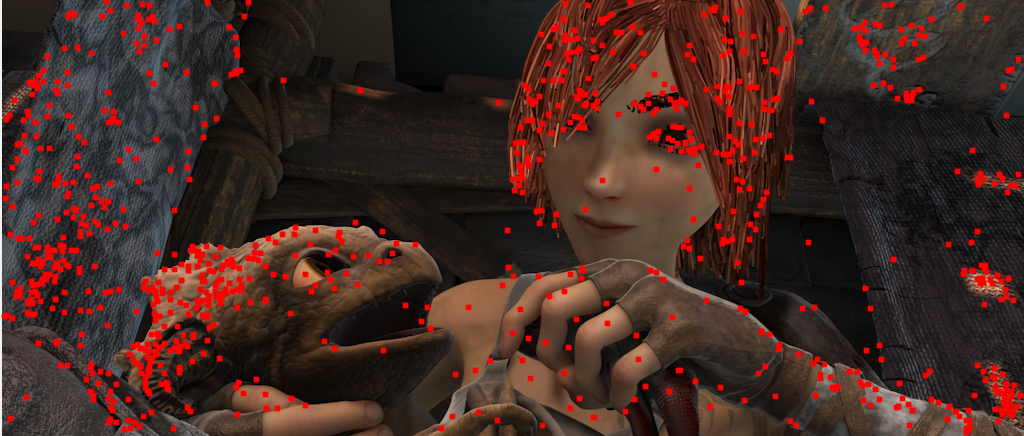}
  \label{fig:alley_1_F_20}}
  \subfloat[Second frame]{
    \includegraphics[width=0.3\textwidth]{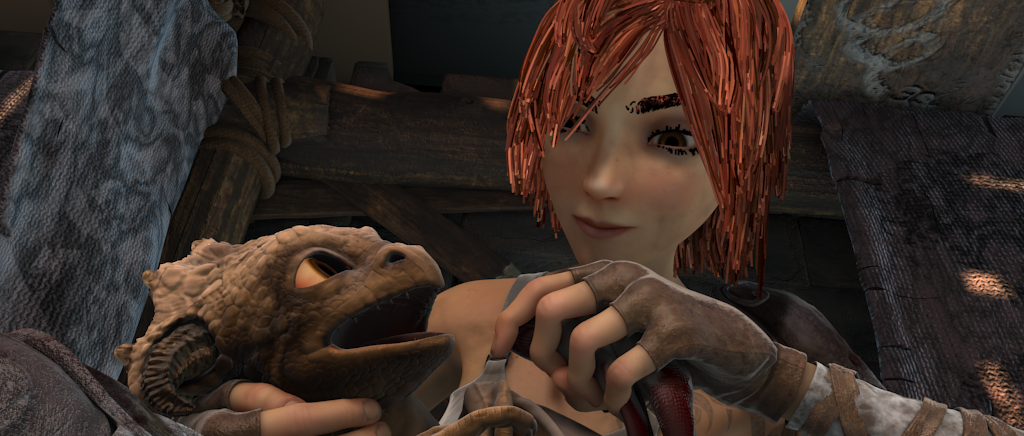}
  \label{fig:alley_1_F_21}}
  \subfloat[Ground truth]{
    \includegraphics[width=0.3\textwidth]{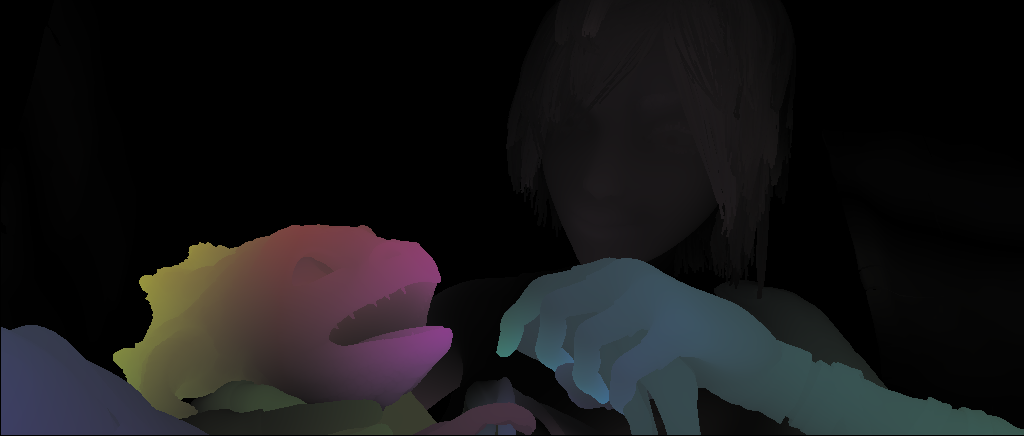}
  \label{fig:alley_1_20_gt}}
  \\
  \subfloat[30\% estimated]{
    \includegraphics[width=0.3\textwidth]{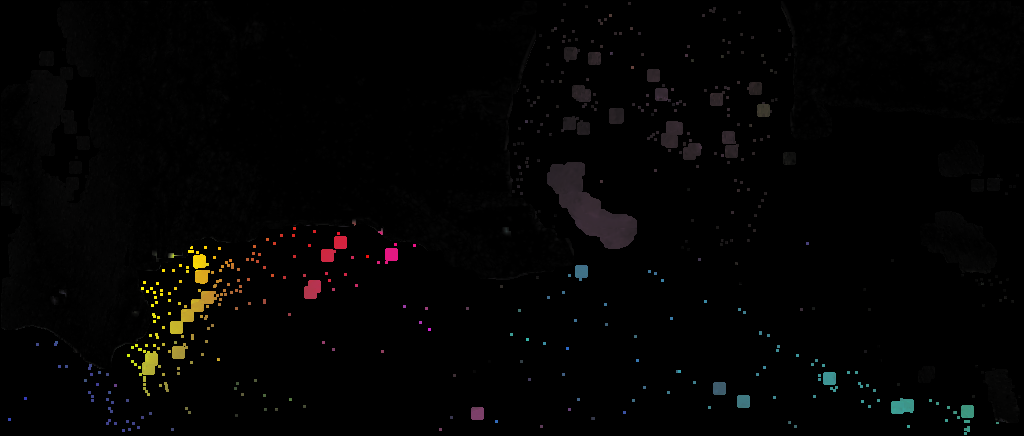}
  \label{fig:alley_30}}
  \subfloat[70\% estimated]{
    \includegraphics[width=0.3\textwidth]{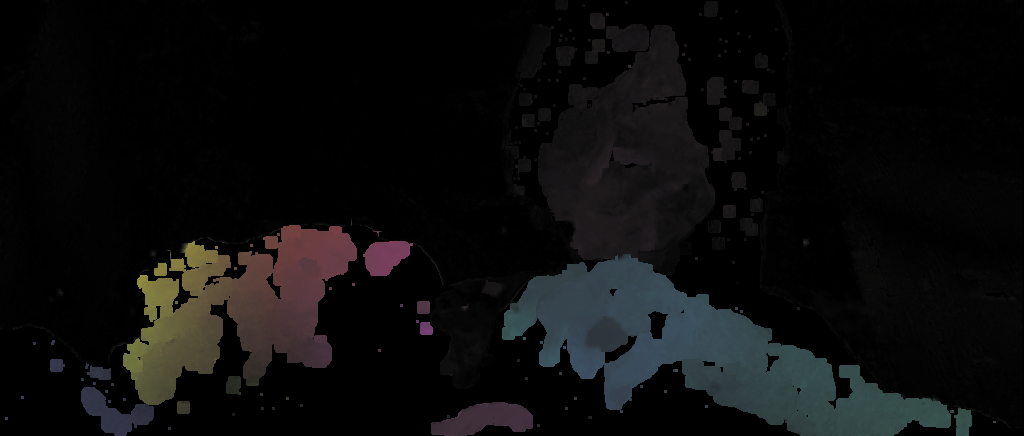}
  \label{fig:alley_70}}
  \subfloat[80\% estimated]{
    \includegraphics[width=0.3\textwidth]{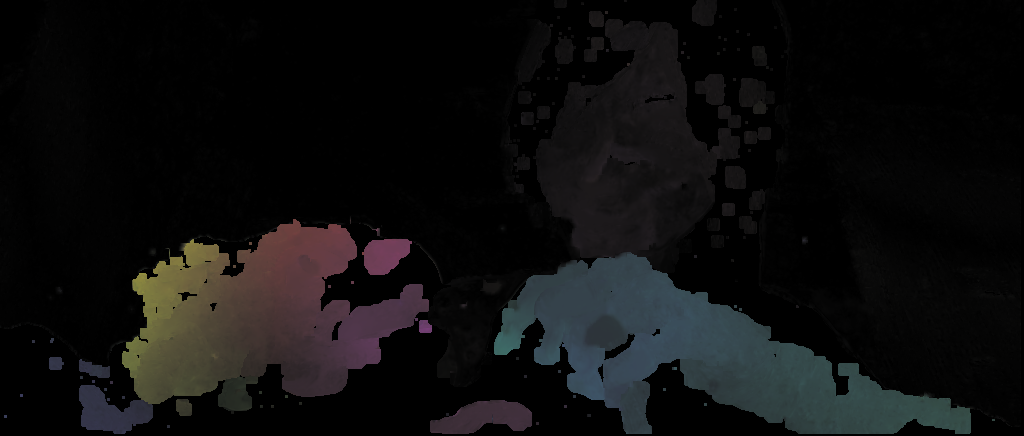}
   \label{fig:alley_80}}
  \\
  \subfloat[95\% estimated]{
    \includegraphics[width=0.3\textwidth]{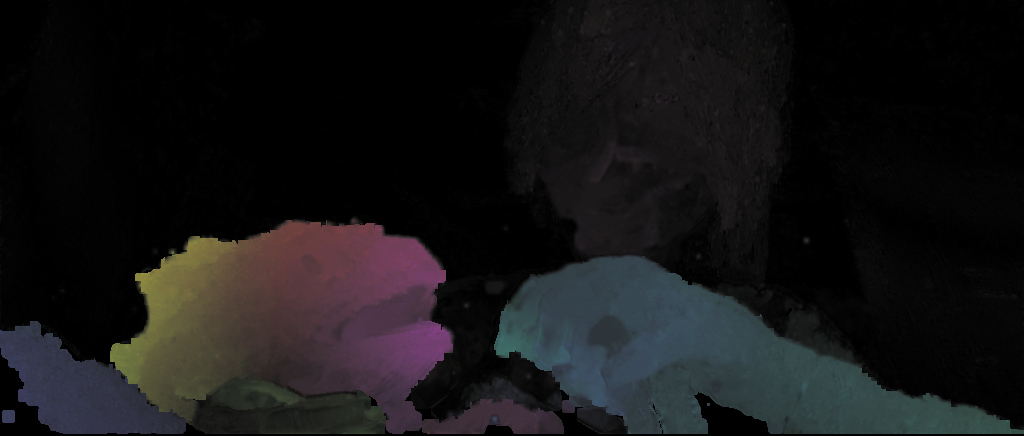}
  \label{fig:alley_95}}
  \subfloat[100\% estimated]{
    \includegraphics[width=0.3\textwidth]{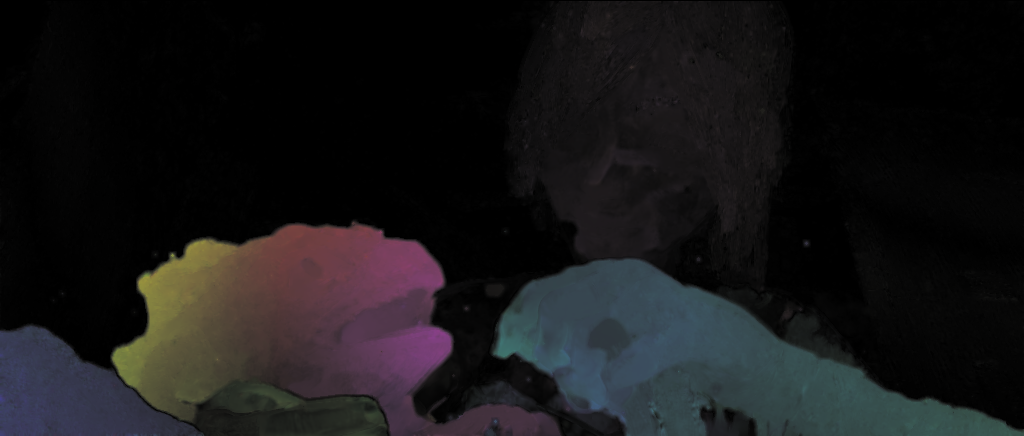}
  \label{fig:alley_100}}
  \subfloat[Final]{
    \includegraphics[width=0.3\textwidth]{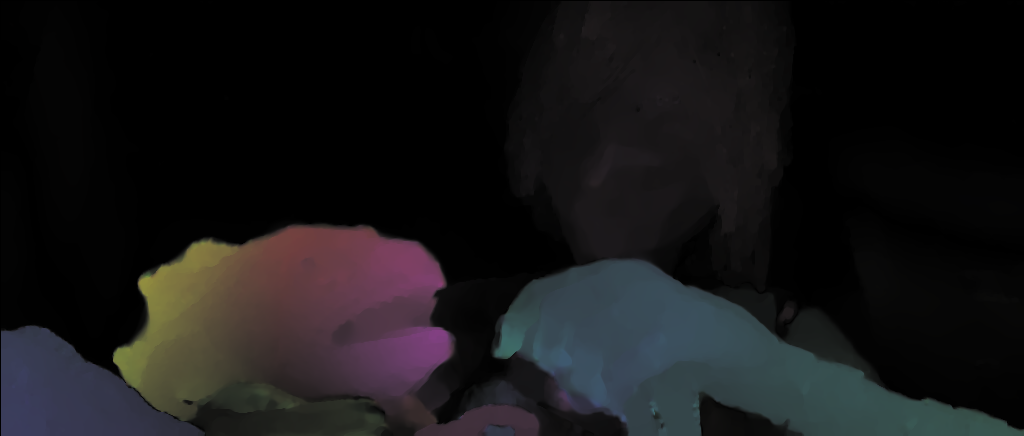}
   \label{fig:alley_var}}
  \caption{Example of a sparse-to-dense evolution of the estimated optical flow during the local step, starting from a set of initial seeds provided by SIFT (shown in red in (a)). The final estimate  (after the global minimization step) is also shown. The optical flow values are represented using the color coding scheme shown in Fig.~\ref{fig:sintel_claim_alley1}. The energy functional used is the classical $TV_{\ell_2}$-$L1$.}\label{fig:evolution_example}
\end{figure*}

Energy-based methods are called global methods since they find correspondences by minimizing an energy defined on the whole image. 
Traditionally, global methods include a linearization of the warped image in the data term to make the optimization problem more tractable. The linear approximation is only valid for small displacements, that is why these methods are usually embedded in a coarse-to-fine warping scheme in order to better capture large displacements. However, they still fail to correctly handle large motions of small objects (not present at coarser scales). On the other hand, local methods are much better adapted to capture large movements. 
The work of \'Alvarez~\etal~\cite{alvarez2000reliable} was probably the first to note that the standard coarse-to-fine approach may not be enough and proposed to modify it by using a linear scale-space focusing strategy from coarse to fine scales in order to avoid convergence to incorrect local minima. More recently, Steinbr{\"u}cker~\etal~\cite{steinbrucker2009large} proposed an algorithm that does not require the coarse-to-fine warping strategy. It was one of the first attempts to capture large motions with variational methods, where the data term and the regularizer are decoupled by a quadratic relaxation technique and the optimization problem is directly solved at the finest scale by alternating two global minimizations while decreasing the decoupling parameter. One of the minimization problems is convex; the other one is non-convex but it is a point-wise optimization problem, so its solution is found using an exhaustive search for every pixel, which can be performed in parallel. 
Another early work is the proposal by Brox and Malik~\cite{Brox_Malik_LD}, that incorporates sparse matches into the variational model by adding an extra term that penalizes deviations of the estimated optical flow from the flow given by the matches obtained by HOG descriptors \cite{dalal2005histograms} at some sparse locations. The extended variational method is solved with a coarse-to-fine strategy as usual.
DeepFlow \cite{Weinzaepfel2013} and SparseFlow \cite{timofte2015sparse} follow the same strategy as proposed by Brox and Malik~\cite{Brox_Malik_LD}, the difference among them lying in the way they find  the matches that define the matching term in the variational approach. The descriptor matching proposed by Weinzaepfel~\etal~\cite{Weinzaepfel2013} is inspired by non-rigid 2D warping and deep convolutional networks and permits a more dense and deformable matching compared to the popular HOG/SIFT-like descriptors \cite{dalal2005histograms,SIFT}. The matching algorithm proposed by Timofte~\etal~\cite{timofte2015sparse} is also robust to non-rigid deformations and is based on the compressed sensing theory.
The proposal by Xu~\etal~\cite{xu2012motion} also  uses a coarse-to-fine approach but the initial flow in each level of the pyramid is modified. At each level, different candidate flows are considered: the flow propagated from the previous level, sparse feature matching (computed by SIFT), and dense nearest neighbor patch matching. Then, for each pixel, the optimal flow is selected among the different candidates by solving a discrete optimization problem.

Other works that do not use a coarse-to-fine strategy are based on correspondences obtained by a nearest neighbor algorithm; one is based on a purely local method  \cite{bao2014fast} while another
\cite{chen2013large} refines the initial dense (and noisy) correspondence field using a motion segmentation and the minimization of a global energy. 
%
Some recent methods are based on a  sparse-to-dense approach; they
 start from a sparse set of correspondences that capture large displacements and these are densified  by edge-aware interpolation techniques  \cite{leordeanu2013locally,EpicFlow}. Later, the densified flow is refined by minimizing a global energy at the finest scale.
Other works \cite{menze2015discrete,yang2015dense} also include a final refinement of the flow by minimizing a global continuous 
energy in the original image scale directly; either using a discrete inference problem in a conditional random field as a first step~\cite{menze2015discrete}, or a piecewise adaptive parametric model embedded in an energy function that combines both continuous and discrete variables~\cite{yang2015dense}.
The recent proposal of Fortun~\etal~\cite{fortun2014aggregation} solves a discrete optimization of a global energy in the original scale where the different flow candidates are obtained in a previous step by 
patch-based parametric motion estimation; this method also makes a special treatment of occlusions.  
On the other hand, an almost purely data-based dense optical flow is indeed possible, by working with approximate nearest neighbor fields with a hierarchical search strategy and an advanced outlier filtering~\cite{bailer2015flowICCV}.

\begin{figure*}[htbp]
\centering
\subfloat[First frame]{
    \includegraphics[width=0.3\textwidth]{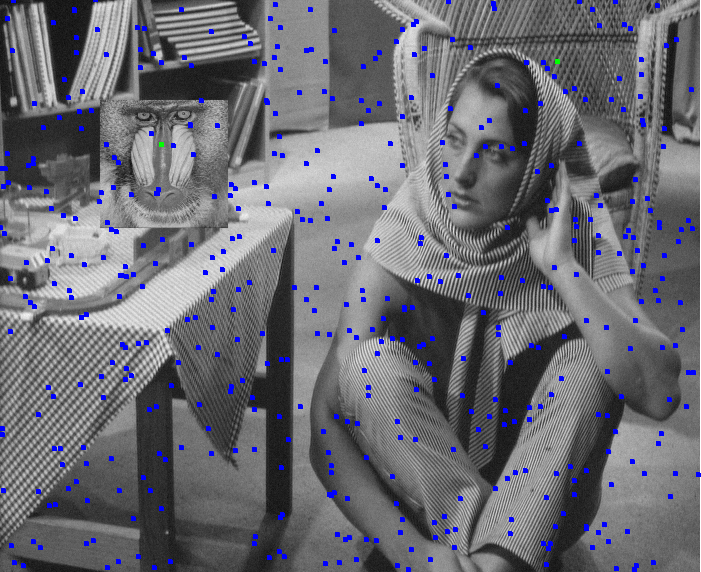}
  \label{fig:abruta}}
  \subfloat[Second frame]{
    \includegraphics[width=0.3\textwidth]{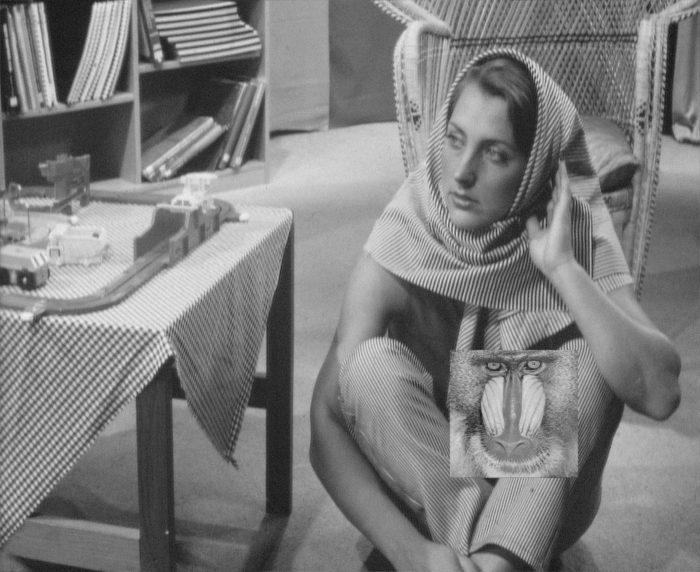}
  \label{fig:bbruta}}
  \subfloat[Ground truth]{
    \includegraphics[width=0.3\textwidth]{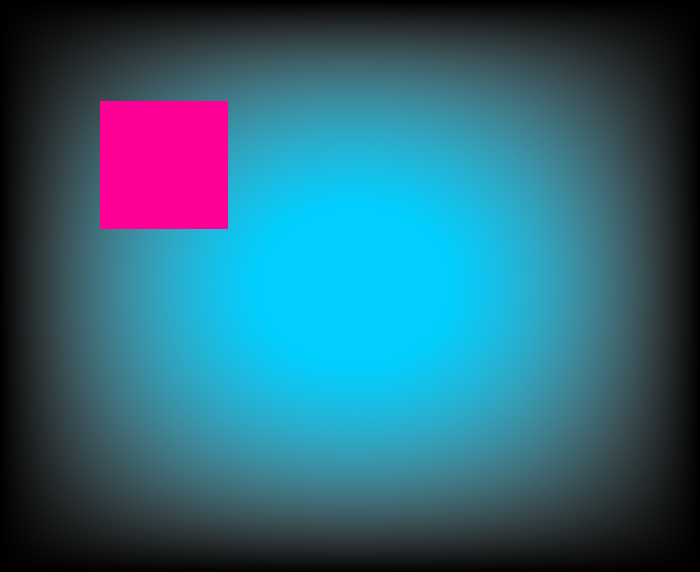}
  \label{fig:gt_macaco}}
  \\
  \subfloat[10\% estimated]{
    \includegraphics[width=0.3\textwidth]{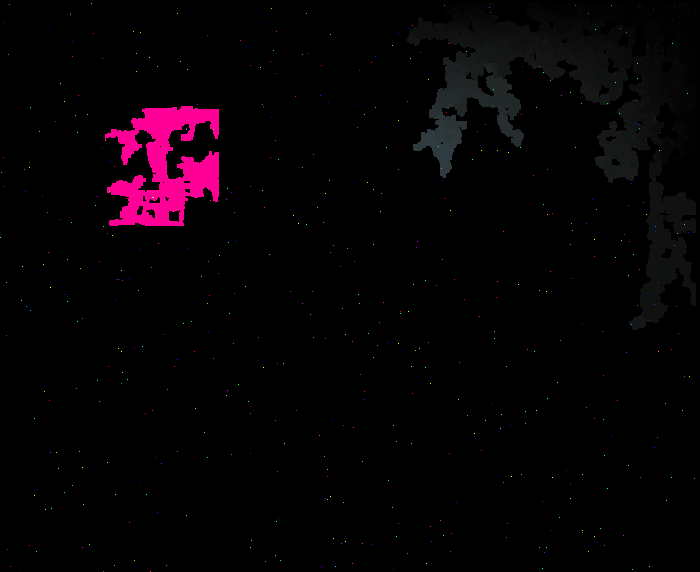}
    \label{fig:macaco_00}}
  \subfloat[30\% estimated]{
    \includegraphics[width=0.3\textwidth]{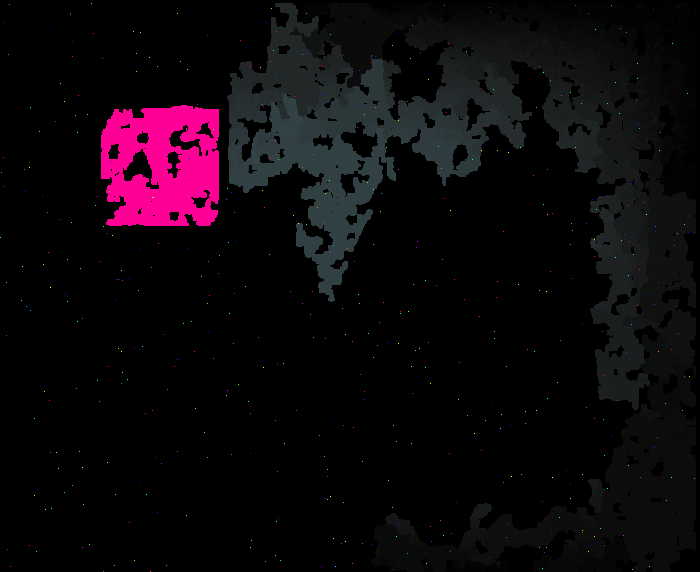}
    \label{fig:macaco_30}}
  \subfloat[50\% estimated]{
    \includegraphics[width=0.3\textwidth]{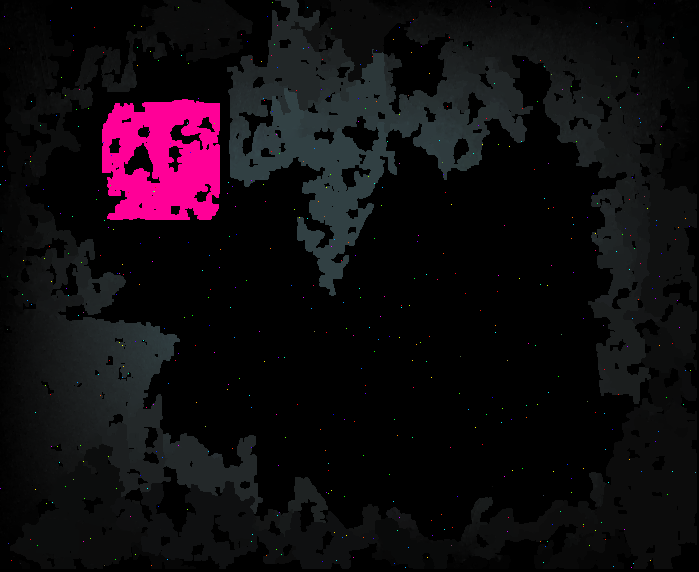}
    \label{fig:macaco_50}}
    \\
  \subfloat[70\% estimated]{
    \includegraphics[width=0.3\textwidth]{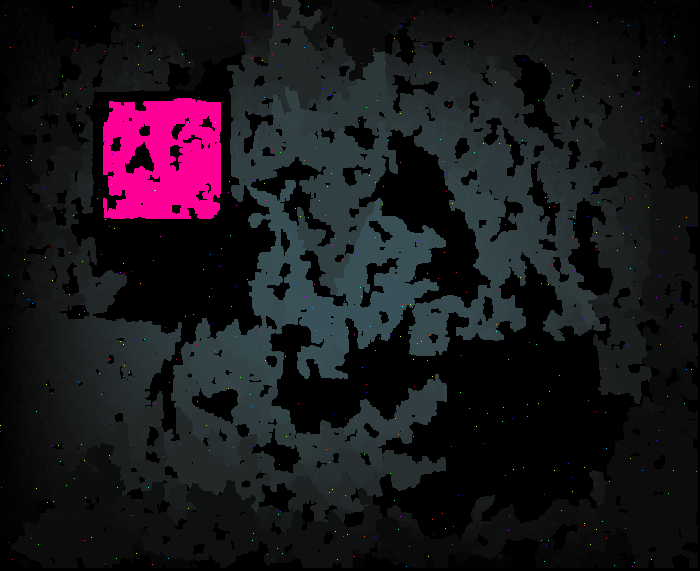}
    \label{fig:macaco_70}}
  \subfloat[100\% estimated]{
    \includegraphics[width=0.3\textwidth]{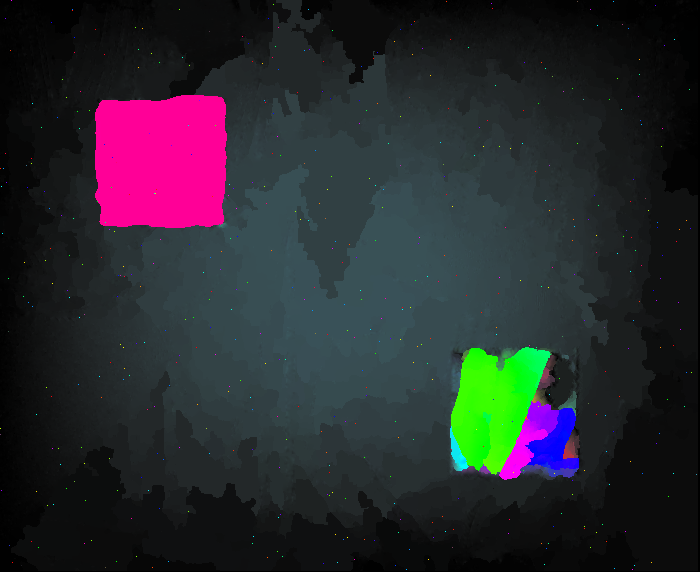}
    \label{fig:macaco_100}}
  \subfloat[Final]{
    \includegraphics[width=0.3\textwidth]{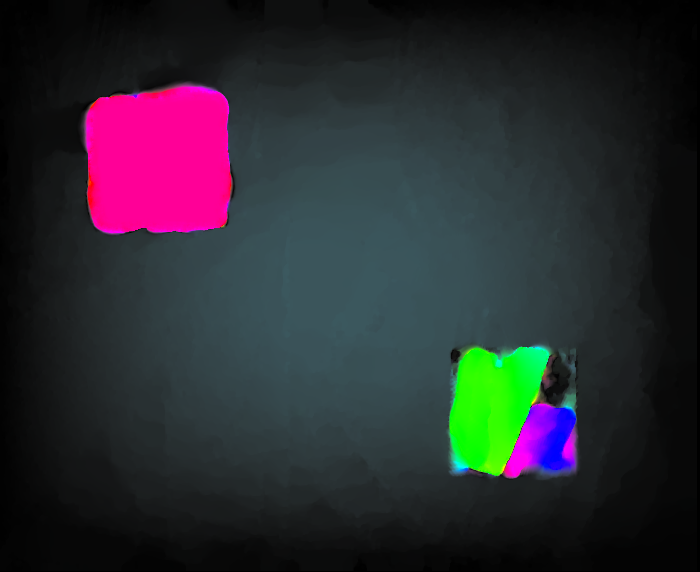}
    \label{fig:macaco_var}}
  \caption{Sparse-to-dense evolution of the fixed coordinates for a toy sequence made of the composition of a a large displacement (moving the \textit{Baboon} small image) and a smooth movement of the background. The initial set of sparse matches contains only 2 correct matches (obtained by SIFT and shown in green) and  508 outliers (shown in blue). Even in the presence of a huge number of outliers, the method recovers the correct motion in all the image with the exception of the occluded region since the chosen functional ($TV_{\ell_2}$-$L1$) does not take into account occlusions. For visualization purposes and better observe the sweep pattern in the optical flow estimation, in this figure we have increased the color contrast. The optical flow values are represented using the color coding scheme shown in Fig.~\ref{fig:sintel_claim_alley1}.}\label{fig:toy_outliers_sample}
\end{figure*}

Here, we also propose to get rid of the coarse-to-fine strategy and directly minimize the energy at the finest scale with the help of some sparse correspondences that capture the large displacement motions. The main difference with respect to the previous works~\cite{leordeanu2013locally,EpicFlow} is the way of combining the sparse matches with the variational approach: the initial matches are densified by a region growing process driven by the minimization of a local version of the target energy functional. Thus, the same variational tool is used both for the densification step and for the final global optimization problem. 
In contrast, the sparse-to-dense interpolation of the sparse matches in \cite{leordeanu2013locally,EpicFlow} is based on a regularity measure that takes into account occlusion and boundary constraints. 
In this way, since in our case we are always taking into account a data term and a regularizer, together with a smart local minimization of the energy guided by the best matches, our sparse-to-dense intermediate step is more robust to outliers in the initial set of sparse correspondences and is able to correctly recover a dense smooth motion in a certain region with just a single correspondence in it and without the need of using edge information (see Figures \ref{fig:evolution_example} and \ref{fig:toy_outliers_sample}). The outliers in this experiment were randomly introduced to test our claim against a huge percentage of them. Nevertheless, the robustness of our iterated algorithm to usual outliers is shown in Figure~\ref{fig:its_local_minimization}, the different Tables and remaining experiments and it relies on pruning strategies based on saliency and forward-backward optical flow consistency. 

\section{Proposed minimization strategy}
\label{sec:faldoi}

In this section we present a minimization strategy that can be applied to any optical flow energy functional and which is founded on estimating a good local minimum with the help of a discrete set of matches. It is able to benefit both from the sparse techniques, which handle arbitrarily large displacements, and from the continuous optimization of a variational formulation, which yields dense flow fields with subpixel accuracy.  The basic idea of the method is to assume that at least some matches are correct, and propagate the correct information from those seeds driven by the minimimization of the energy around them. This is ilustrated in Figure~\ref{fig:sintel_claim_alley1}, where each region of smooth movement has at least one correct seed.

The sparse set of initial correspondences (we will refer to them as \textit{seeds}) 
is used as a reference or guide to recover a dense flow field. This is done by iteratively growing the seeds by a local (patch-based) minimization of the functional in a proper order 
(detailed in Sect.~\ref{ssec:local_min} and Sect.~\ref{ssec:iteratedfaldoi}). 
This dense optical flow is then refined by a global minimization of the energy. The algorithm always works at the finest scale of the image. A pseudo-code of the whole strategy is presented in Algorithm~\ref{algo:faldoi} while Algorithm~\ref{algo:iterated-faldoi} presents a slightly modified version which is explained in Sect.~\ref{ssec:iteratedfaldoi}.

In order to detail the proposed approach, we  introduce some notation and assumptions. 
Let us denote two consecutive image frames of a video sequence as $I_t, I_{t+1}:\Omega\to R$. As usual, we assume that the image domain $\Omega$ 
is a rectangle in $ R^2$. In order to compute the optical flow $\mathbf{u}:\Omega\to R^2$ between $I_t$ and $I_{t+1}$, we use a discrete set of matches ${\cal{M}}=\{(\mathbf{x}_i,\mathbf{y}_i)\}$, $i=1,\ldots,N$, and an energy $E(\mathbf{u})$, defined from  $I_t$ and $I_{t+1}$. Minimizing $E$ on an appropriate set, a minimum $\mathbf{u}$ of $E$ represents an optical flow between $I_t$ and $I_{t+1}$.  
We assume that the discrete matches in ${\cal{M}}$ have been computed with a sparse matching algorithm in such a way that $\mathbf{x}_i$ is thought as belonging to the image domain of $I_t$ and  $\mathbf{y}_i$ as belonging to the image domain of $I_{t+1}$. From these discrete matches, we compute the initial set of seeds, denoted here by $S$, by defining $\mathbf{u}(\mathbf{x}_i)=\mathbf{y}_i-\mathbf{x}_i$,  $i=1,\ldots,N$.
Each seed $\mathbf{p}$ in the finite set of initial seeds $S$ stores the corresponding pixel $\mathbf{x}^{\mathbf{p}}$, its related optical flow $\mathbf{u}^{\mathbf{p}}$ and the local energy of the functional around it (see Algorithm~\ref{algo:faldoi}). 


\subsection{Computing a good local minimum}

This step  can be interpreted as an adaptive grouped coordinated descent approach driven by the lowest local values of the energy on patches centered at pixels. It is inspired from  the match propagation principle \cite{MatchPropag}, where a set of initial sparse matches, the seeds, are propagated to neighboring pixels using a similar technique to the region growing strategy. This principle was used in the work of \cite{Kannala_Brandt} in order to obtain a quasi-dense disparity map, assuming that the seeds and their neighboring pixels may present similar disparity values. Moreover, it shares ideas with the coordinate descent methods, which are optimization techniques in multi-variable functions. They minimize the objective function solving a sequence of one variable minimization problems. Each subproblem improves the estimate of the solution by minimizing along a selected coordinate while all other coordinates are fixed. Generally, each coordinate is visited several times to reach a minimum. The sweep pattern is the name used to define how the alternating minimization process moves along all the coordinates. If there is a fixed order to visit the coordinates, this is called path-wise coordinate descent \cite{Frie_Tibs} or cyclic coordinate. In our case, the election of the sweep pattern has a fundamental role during the minimization process; we will follow an adaptive choice of the sweep pattern driven by a seed growing algorithm based on the value of the local minimization of the energy in a patch centered at that coordinate (or pixel in our case).
The sweep process is managed through a priority queue where the potential candidates (optical flow) for each coordinate are inserted. 
Each candidate presents a related energy that is used to determine its position in the queue: Candidates with less associated local energy will be at the top of the priority queue. The initial seeds are inserted with zero energy.

In the following we will first present the baseline algorithm for the local minimization step, 
where every pixel is visited just once. Then we explain how it is iterated in order to revisit the pixels several times  and gain more robustness against occlusions and resistant outliers of the sparse matcher. 

\subsubsection{Baseline algorithm: faldoi}
\label{ssec:local_min}

Initially, the seeds are inserted to the priority queue (containing, as said before, the value of the optical flow and  the energy on a local patch centered at the pixels) with zero energy. 
Along the minimization process, new candidates will be added to the queue. This collection of potential candidates for each coordinate will be sorted based on their local energies. Whenever an element is removed from the queue to fix a coordinate, we are selecting the candidate  with the lowest associated local energy. 
The aforementioned process is repeated until the priority queue is empty and a dense optical flow is obtained from the initial seeds. There may be several candidates for the same pixel in the queue; when a candidate is extracted from the queue to fix it, if its corresponding pixel has already been fixed (by a candidate with lowest energy) nothing is done. Then, after a certain extractions the queue is emptied.

Each time that a coordinate is extracted from the queue to fix it, a local minimization of the energy $E(u)$ is solved on a square patch centered around the pixel previously fixed. Then, the estimated optical flow values of its neighbors are inserted as potential candidates into the priority queue with the energy (after local minimization) of the patch centered at the fixed coordinate. 

Whenever the energy is minimized in a local patch  ${\cal{P}}$ centered at the coordinate that 
has been fixed we need to provide an initial flow in the unknown values in ${\cal{P}} \cap W$ (where $W$ is the set of locations where the optical flow has not been fixed yet). The initialization is just an interpolation of the already fixed optical flow values in the patch through the Laplace equation resulting from the minimization of the following Dirichlet energy 
    \begin{eqnarray*}
      \min_{u_i} & \int_{\cal{P}} \left|\nabla u_i \right|^{2} d\x  \quad \hbox{\rm s.t.  } & u_i = u^0_i \; \hbox{\rm in } {\cal{P}}  \cap W^{C}
    \end{eqnarray*}  
where $i=1,2$, $\mathbf{u}=(u_1,u_2)$, and $\mathbf{u}^0=(u^0_1,u^0_2)$ contains the optical flow of the already fixed coordinates in ${\cal{P}} \cap W^C$ ($W^C$ denotes the complementary set of $W$). The Euler-Lagrange equation derived from the previous energy is the Laplace equation, which is solved by gradient descent with Neumann boundary conditions. Other intra-patch interpolations are allowed (e.g., based on the bilateral filter or even a constant interpolation based on the flow values on ${\cal{P}}  \cap W^{C}$).

The details of the energy minimization process are given in~Appendix \ref{appendix:optalgos}
and Algorithm \ref{algorithm:global_min}. In practice, we consider patches of $11 \times 11$ pixels (although bigger patches can be used, it increases the computational cost which is proportional to the size of the patch) and we perform 10 iterations of the minimization process, in every local patch (line 7 of Algorithm \ref{algo:basic-faldoi-growing} ), of a version of $E(u)$ where the warped image has been linearized  with a single warping. 

\begin{algorithm}[H]
  \caption{\texttt{faldoi}
    (large displacement optical flow with astute initialization) 
  }\label{algo:faldoi} 
  \DontPrintSemicolon
  \SetKwInOut{Input}{Input}
  \SetKwInOut{Output}{Output}
  \SetKwFunction{flowrefinement}{flow-refinement}
  \SetKwFunction{flowdensification}{flow-densification}
  \SetKwFunction{extractpatch}{extract-patch}
    \SetKwFunction{interpolate}{interpolate}
    \SetKwFunction{basicfaldoigrowing}{basic-faldoi-growing}
  \Input{Images~$A, B$}
  \Input{Functional~$E$}
  \Input{Image matcher~$M$}
  \Input{Patch size~$w$}
  \Output{Flow~$\mathbf{u}$}
  $\mathbf{u}\leftarrow NULL$
  \tcp*[r]{initialize flow field with empty data}
  $Q\leftarrow\emptyset$
  \tcp*[r]{initialize an empty priority queue}
  $(\x_i,\y_i)_{i=1,\ldots,N}\leftarrow M(A,B)$
  \tcp*[r]{compute discrete matches}
  \For(\tcp*[f]{add the matches as seeds with zero energy}){$i=1,\ldots N$}{
    $Q.push(0, \x_i, \y_i-\x_i)$
  }

  $\u\leftarrow\basicfaldoigrowing(A,B,\mathbf{u},E,Q,w)$ \tcp*[r]{basic faldoi growing}
    $\u\leftarrow\flowrefinement(A,B,\Omega,E,\mathbf{u})$
  \tcp*[r]{Minimize~$E_{A,B}(\mathbf{u})$ over $\Omega$ (Algorithm \ref{algorithm:global_min})}
\end{algorithm}

\begin{algorithm}[H]
	\caption{\texttt{basic-faldoi-growing}
	(densify an incomplete flow)
	}\label{algo:basic-faldoi-growing} 
	\DontPrintSemicolon
	\SetKwInOut{Input}{Input}
	\SetKwInOut{Output}{Output}
    \SetKwFunction{flowdensification}{flow-densification}
	\SetKwFunction{extractpatch}{extract-patch}
	\SetKwFunction{flowrefinement}{flow-refinement}
	\SetKwFunction{assert}{assert}
	\SetKwFunction{interpolate}{interpolate}
	\SetKwFunction{interpolateusinglaplaceequation}{interpolate-using-laplace-equation}
	\Input{Images~$A, B$}
	\Input{Flow~$\mathbf{u}_0$}
	\Input{Functional~$E$}
	\Input{Priority queue~$Q$}
	\Input{Patch size~$w$}
	\Output{Flow~$\mathbf{u}$}
 \While{$Q.num\_elements() > 0$}
  {
    $e,\x,\v\leftarrow Q.pop()$
    \tcp*[r]{get the candidate with lowest energy}
    \If{$\mathbf{u}(\x)= NULL$}{
      $\mathbf{u}(\x)\leftarrow\v$
      \tcp*[r]{fix the field for this candidate}
    }
            ${\cal{P}}_{\x} \leftarrow\extractpatch(\Omega,\x,w)$
        \tcp*[r]{extract patch of size~$w\times w$ around $x$}
          $\mathbf{u}\leftarrow\interpolate(\mathbf{u},\cal{P}_{\x})$
  \tcp*[r]{fill-in missing values}
  $\mathbf{u}\leftarrow\flowrefinement(A,B,\cal{P}_{\x},E,\mathbf{u})$
  \tcp*[r]{Minimize~$E_{A,B}(\mathbf{u})$ over $\cal{P}_{\x}$ (Algorithm \ref{algorithm:global_min})}
    $e\leftarrow E_{A,B}(\mathbf{u},\cal{P}_{\x})$
  \tcp*[r]{compute the energy of the solution}
    \For(\tcp*[f]{traverse the neighbors of~$\mathbf{x}$}){$\y\in\mathcal{N}(\x)$}{
      \If{$\mathbf{u}(\y)=NULL$}{
        $Q.push(e,\y,\mathbf{u}(\y))$
        \tcp*[r]{push this candidate value}
      }
    }
  }
\end{algorithm}
\subsubsection{Iterated faldoi} \label{ssec:iteratedfaldoi}
Given the fact that, generally, in coordinate descent methods each coordinate has to be visited several times in order to reach a minimum, we propose to  revisit the coordinates and thus perform several iterations of the  baseline algorithm presented in Sect.~\ref{ssec:local_min} with a forward-backward pruning of the flow values between two consecutive iterations in order to gain more robustness against resistant outliers.

More precisely, between two consecutive iterations of the coordinate descent completing a sweep pattern, we perform a pruning based on a forward-backward consistency check. The goal is to remove wrong matches, specially in occlusion areas and also due to self-similarity. The latter may appear due to outliers in the initial seeds that have expanded by the region growing scheme.  
Our algorithm computes both the forward,  $\mathbf{u}^F$, and the backward,  $\mathbf{u}^B$, optical flows between any two frames $I_t$ and $I_{t+1}$. Then, a forward-backward consistency check of these flows is performed and the forward flow values at points $\mathbf{x}\in\Omega$ not verifying
 $
 \|\mathbf{u}^F(\mathbf{x})+\mathbf{u}^B(\mathbf{x}+\mathbf{u}^F(\mathbf{x}))\|<\epsilon 
$
 are removed, where $\epsilon>0$ is a small constant (it is set to $2$ in our experiments). In the same way, the backward flow at points $\mathbf{x}\in\Omega$ not verifying
$
 \|\mathbf{u}^B(\mathbf{x})+\mathbf{u}^F(\mathbf{x}+\mathbf{u}^B(\mathbf{x}))\|<\epsilon
$
 are removed.

\begin{figure*}
\begin{center}
  \includegraphics[width=0.3\linewidth]{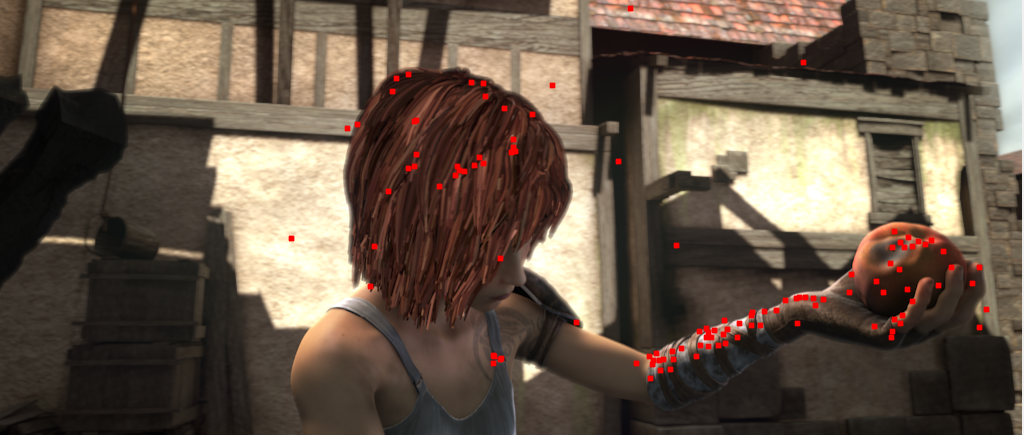}
    \includegraphics[width=0.3\linewidth]{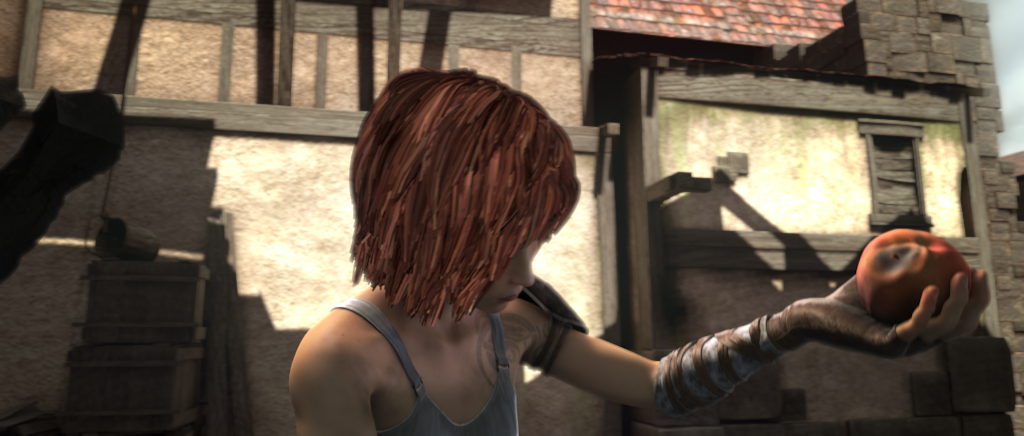}\\
  \includegraphics[width=0.3\linewidth]{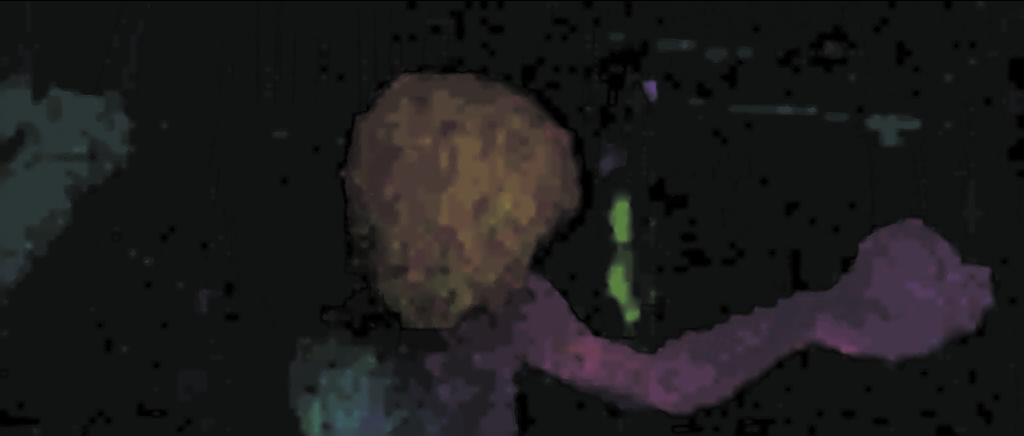}
    \includegraphics[width=0.3\linewidth]{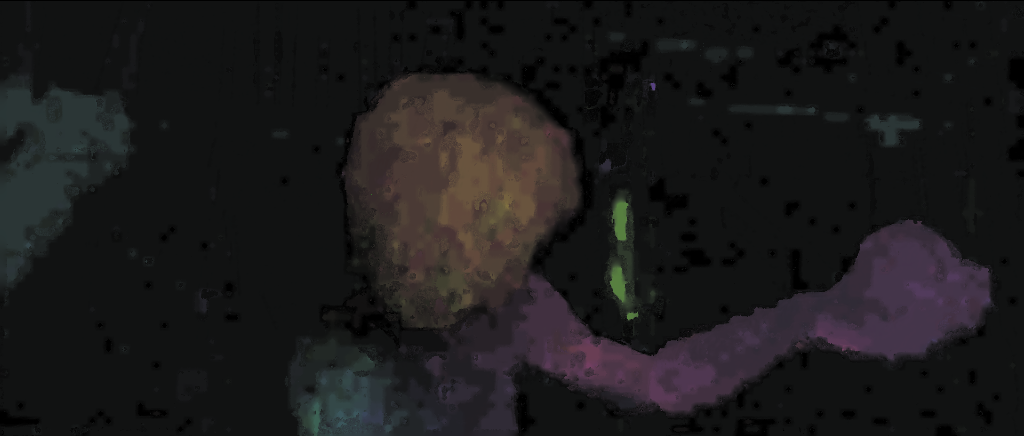}
      \includegraphics[width=0.3\linewidth]{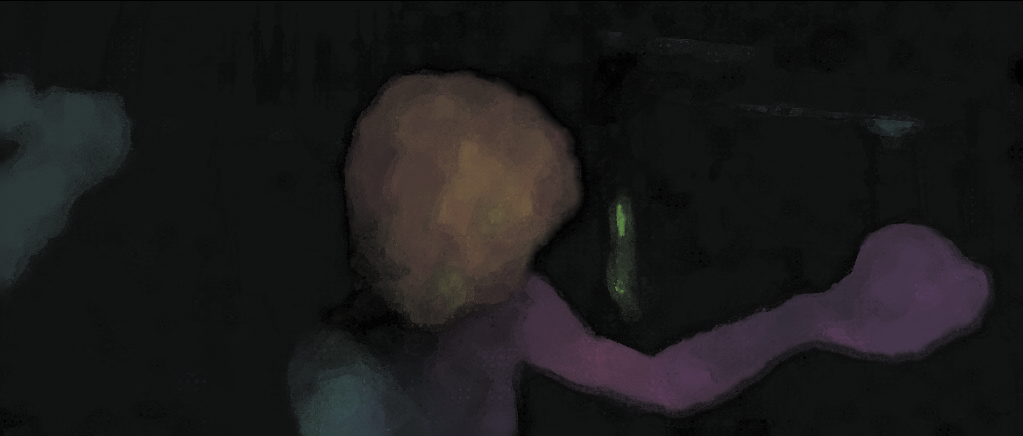}\\
        \includegraphics[width=0.3\linewidth]{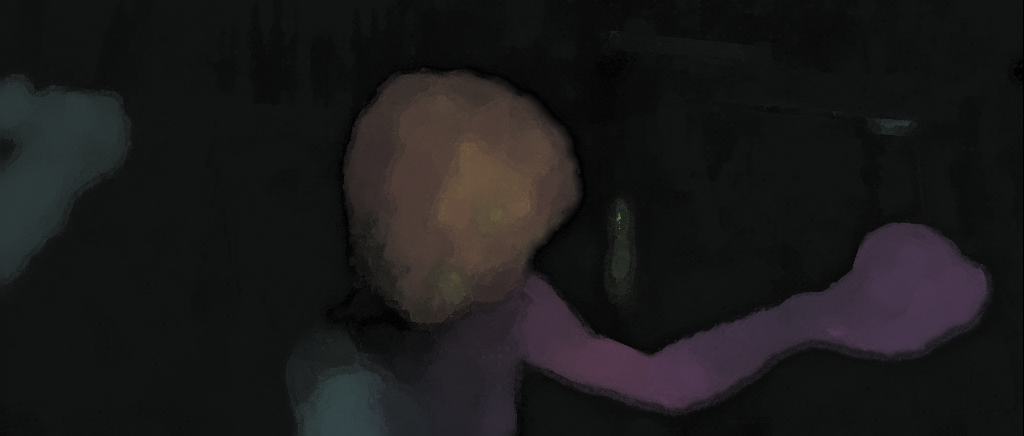}
    \includegraphics[width=0.3\linewidth]{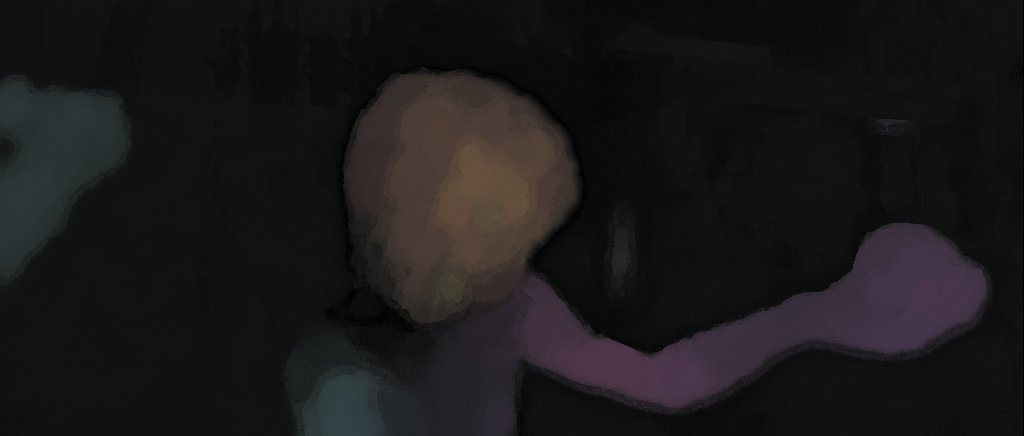}
      \includegraphics[width=0.3\linewidth]{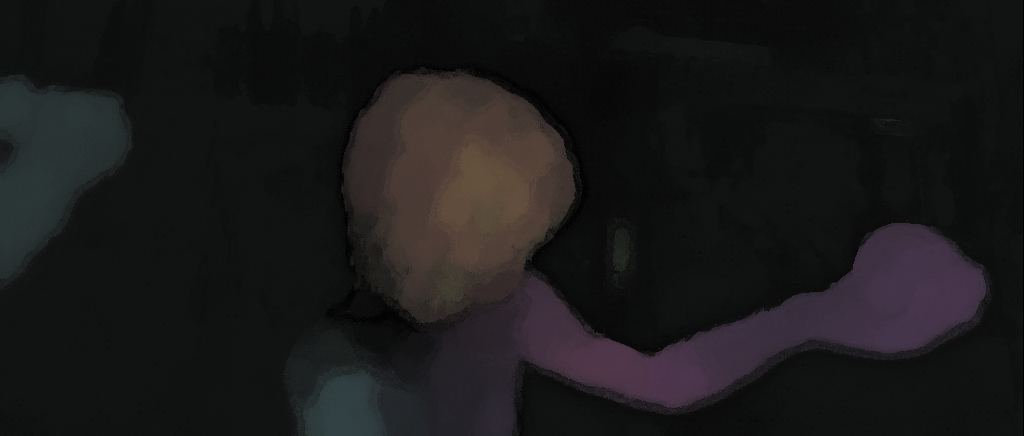}\\
       \includegraphics[width=0.3\linewidth]{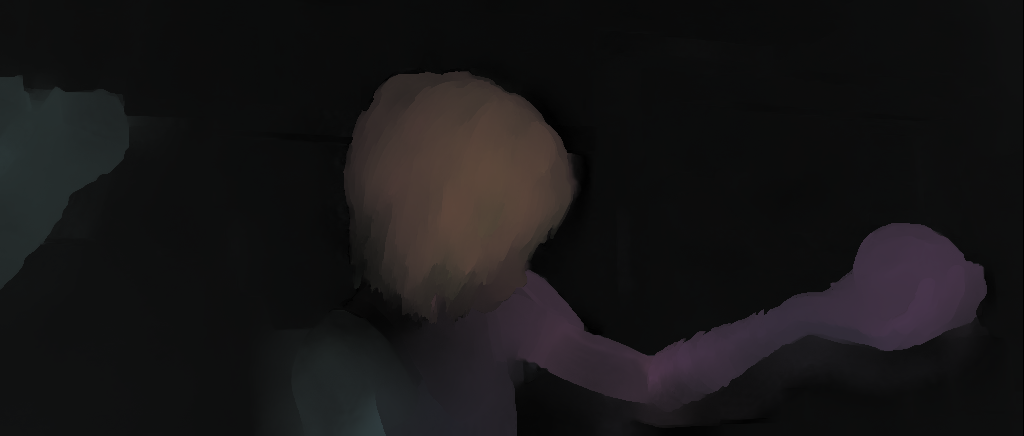}
          \includegraphics[width=0.3\linewidth]{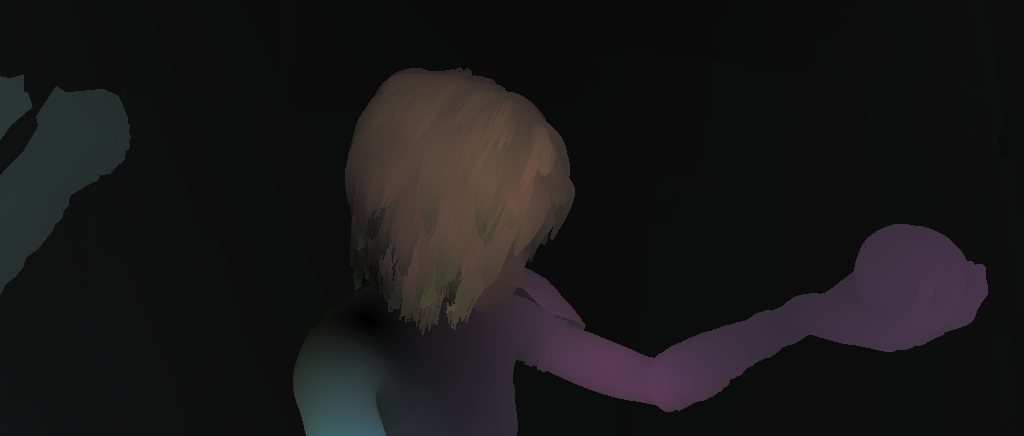}  
\end{center}
   \caption{Iterated faldoi strategy applied to the $NLTV$-$CSAD$ energy corresponding to MAX\_IT=6 in Algorithm~\ref{algo:iterated-faldoi}. The evolution of the optical flow estimate is shown from left to right and top to bottom. The first line shows the two frames with the initial seeds superimposed in red in the first frame. The second and third lines display the six iterations of the sweep pattern in the local minimization step. The last line shows the final estimate (global minimization) and the ground truth. Let us notice how the motion boundaries are progressively refined, the piecewise constant appearance disappears, and the outlier is considerably  reduced, thus adding robustness to outliers. In this experiment the seeds have been computed using Deep Matching, which gives more outliers than SIFT (used for the experiment with the same pair of images in Figure~\ref{fig:evolution_example}).}
\label{fig:its_local_minimization}
\end{figure*}


In the iterated faldoi, the sweeps following the first one are slightly different from the baseline sweep of Sect.~\ref{ssec:local_min} because they 
start not from a sparse set of optical flow values but from a dense motion field with holes that arise after the forward-backward pruning. The two main differences with respect to the main sweep are:

\noindent
1. \textit{Initialization of the queue.} The seeds that survive to the forward-backward consistency check are inserted into a new priority queue with zero local energy. The rest of optical flow values that also survive to 
the forward-backward pruning are added with its associated local energy.

\noindent
2. \textit{Initialization of the optical flow of a patch before local minimization.} Every time a local patch is minimized an initial flow in the patch is needed. We make a distinction between the pixels that passed the forward-backward consistency test and those that didn't. In the pixels that passed the test the initial flow is the most updated flow value by the last local minimization in that pixel. For that, we use an auxiliary flow which is updated after every local minimization. On the other hand, in the pixels that didn't pass the test we fill-in the initial flow by iterating a bilateral filtering in the local patch.  This intra-patch initialization could be also used in the first baseline sweep of Sect.~\ref{ssec:local_min}. Nevetherless, in that very first interpolation, for efficiency reasons we opted for the rougher Laplace interpolation explained above.

A pseudo-code of the iterated faldoi strategy is presented in Algorithm~\ref{algo:iterated-faldoi}. 
Figure \ref{fig:its_local_minimization}  shows the importance of the iterations of the sweep pattern in the local minimization step. The optical flow is improved from one iteration to the following, in particular, the motion boundaries are progressively better recovered and the effect of the piecewise constant flow is lost. Moreover, this iterated faldoi strategy adds robustness to outliers present in the initial seeds due to the  forward-backward consistency check between two consecutive iterations.
The initial outlier regions are reduced, after six iterations to a single region, which is completely lost after the global iteration step.
In this experiment  the seeds have been computed using Deep Matching, which gives more outliers than SIFT (which was used for the experiment with the same pair of images in Figure~\ref{fig:evolution_example}).  Let us remark that in this example, we did not include the initial pruning of seeds based on the low saliency of patches that is used, e.g., in~\cite{EpicFlow}.

\subsection{Global Minimization}
\label{ssec:global_min}
The last step is a global minimization of $E(u)$, taking as initial condition the dense solution provided by the previous step, the local minimization guided by the initial set of matches. Let us remark that we only minimize the global energy at the finest scale, that is, we avoid the multi-scale approach. During  both steps, local and global, we minimize the same energy functional following the same numerical scheme, which is detailed in~Algorithm \ref{algorithm:global_min} and Appendix \ref{appendix:optalgos}. For the global minimization we perform 4 warpings of the energy.

\begin{algorithm}[H]
	\caption{\texttt{iterated-faldoi}
		(iterated version of faldoi + pruning of wrong matches)
	}\label{algo:iterated-faldoi} 
	\DontPrintSemicolon
	\SetKwInOut{Input}{Input}
	\SetKwInOut{Output}{Output}
	\SetKwFunction{flowrefinement}{flow-refinement}
	\SetKwFunction{flowdensification}{flow-densification}
	\SetKwFunction{basicfaldoigrowing}{basic-faldoi-growing}
	\SetKwFunction{extractpatch}{extract-patch}
    \SetKwFunction{fbpruning}{fb-pruning}
    \SetKwFunction{saliencypruning}{saliency-pruning}
	\Input{Images~$A, B$}
	\Input{Functional~$E$}
	\Input{Image matcher~$M$}
	\Input{Patch size~$w$}
	\Output{Flow~$\mathbf{u}$}
	$\mathbf{u}^F\leftarrow NULL$
	\tcp*[r]{initialize forward flow field with empty data}
	$\mathbf{u}^B\leftarrow NULL$
	\tcp*[r]{initialize backward flow field with empty data}
	$Q^F\leftarrow\emptyset, Q^B\leftarrow\emptyset$
	\tcp*[r]{initialize empty priority queues}
	$(\x^F_i,\y^F_i)_{i=1,\ldots,N}\leftarrow M(A,B)$
	\tcp*[r]{compute discrete forward matches}
		$(\x^B_i,\y^B_i)_{i=1,\ldots,M}\leftarrow M(B,A)$
	\tcp*[r]{compute discrete backward matches}
	$(\x^F_i,\y^F_i)_{i=1,\ldots,N'}\leftarrow \saliencypruning(A,(\x^F_i,\y^F_i)_{i=1,\ldots,N})$
	\tcp*[r]{remove matches in}
		$(\x^B_i,\y^B_i)_{i=1,\ldots,M'}\leftarrow \saliencypruning(B,(\x^B_i,\y^B_i)_{i=1,\ldots,M})$
	\tcp*[r]{flat areas}
	\For(\tcp*[f]{add the matches as seeds with zero energy}){$i=1,\ldots N'$}{
		$Q^F.push(0, \x^F_i, \y^F_i-\x^F_i)$
	}
		\For(\tcp*[f]{add the matches as seeds with zero energy}){$i=1,\ldots M'$}{
		$Q^B.push(0, \x^B_i, \y^B_i-\x^B_i)$
	}
	$n\leftarrow$ MAX-IT
	\tcp*[r]{pre-specified number of iterations}
	\While{$n > 0 $}
	{
		$\mathbf{u}^F\leftarrow\basicfaldoigrowing(A,B,\mathbf{u}^F,E,Q^F,w)$
		\tcp*[r]{basic faldoi growing}
		$\mathbf{u}^B\leftarrow\basicfaldoigrowing(B,A,\mathbf{u}^B,E,Q^B,w)$
		\tcp*[r]{basic faldoi growing}	
		$(\mathbf{u}^F,\mathbf{u}^B)\leftarrow\fbpruning(\mathbf{u}^F,\mathbf{u}^B)$
		\tcp*[r]{forward-backward consistency check }
		\For(\tcp*[f]{add the forward canditates with their energies}){each $\mathbf{u}_i^F \in \mathbf{u}^F$}{
		 $e\leftarrow E_{A,B}(\mathbf{u}_i^F,{\cal{P}}_{\x_i^F})$\\
		$Q^F.push(e, \x^F_i, \u_i^F)$
	    }
	    \For(\tcp*[f]{add the backward canditates with their energies}){each $\u_i^B \in \u^B$}{
	    $e\leftarrow E_{A,B}(\mathbf{u}_i^B,{\cal{P}}_{\x_i^B})$\\
		$Q^B.push(e, \x^B_i, \u_i^B)$
	    }
		$n\leftarrow n - 1$
	}
    $\u \leftarrow\flowrefinement(A,B,\Omega,E,\u^F)$
	\tcp*[r]{Minimize~$E_{A,B}(\u)$ over $\Omega$ (Algorithm \ref{algorithm:global_min})}
\end{algorithm}
\section{Results}\label{sec:results}
Our proposal assumes that two ingredients are given: a discrete set of correspondences between two frames of a video (the seeds) and an  optical flow energy functional, namely, $E(\mathbf{u})$. Section~\ref{ssec:severalEnergies} presents the different energies we have used in this paper and   Section~\ref{ssec:ini_seeds} briefly discusses several possibilities for the initial sparse correspondences. Later on, in Sect.~\ref{sec:experiments} we provide a quantitative and qualitative comparison among the several possibilities for the energy terms, as well as a comparison between our method and several state-of-the-art methods, including methods based on the combination of sparse matches and variational techniques \cite{Brox_Malik_LD,leordeanu2013locally,EpicFlow,Weinzaepfel2013}.
\subsection{Different possibilities for the energy}\label{ssec:severalEnergies}
The proposed framework is independent from the energy functional $E(\mathbf{u})$ and we validate it by using several energies. Following most of the optical flow variational approaches in the literature, the different possibilities will share the common feature of being made of two terms: a data fidelity term $E_D(\mathbf{u})$, and a regularization term $E_R(\mathbf{u})$,
\begin{eqnarray}\label{eq:energy}
   E(\mathbf{u}) = E_{D} (\mathbf{u})  + \beta E_{R} (\mathbf{u}) ,
\end{eqnarray}
where $\mathbf{u}:\Omega\to R^2$, $\Omega$ is the image domain,  $I_{t}, I_{t+1}:\Omega\to R^d$ are two consecutive frames ($d=1$ for gray level images and $d=3$ for color images), and~$\mathbf{u}=(u_1,u_2)$ represents the motion field between them.

Let us also remark that our method allows the inclusion of other terms as, \eg, a third term dealing with the occlusions,  as in~\cite{BGLC,ayvaci2012sparse}, which will not be considered here for the sake of simplicity in the presentation.

With the aim of having a robust data cost, specially under illumination changes, we use a convex and continuous approximation of the data cost based on the Census transform  \cite{hafner2013census,stein2004efficient,zabih1994non},  approximated by a sum of centralized absolute differences,  denoted as $CSAD$ \cite{vogel2013evaluation}, and defined as:
\begin{eqnarray} \label{eq:ED}
 E^1_D(\mathbf{u})  =  \int_\Omega \int_\Omega  | I_t(\mathbf{x}) - I_t(\mathbf{y}) - I_{t+1}(\mathbf{x} + \mathbf{u}) + I_{t+1}(\mathbf{y} + \mathbf{u}) | \chi (\mathbf{x}-\mathbf{y}) d\mathbf{y} d\mathbf{x},
\end{eqnarray}
where  $\chi$ denotes the characteristic function of a square of size $P \times P$ centered at the origin ($P=7$ in our experiments, as in~\cite{vogel2013evaluation}).

On the other hand, 
we also consider the classical point-wise $L^1$ data term that imposes the well-known brightness constancy assumption, namely,
\begin{eqnarray}\label{eq:ED_L1}
E^2_{D} (\mathbf{u}) = \int_{\Omega} \left| I_{t+1}(\mathbf{x}+\mathbf{u}) - I_{t}(\mathbf{x})\right| d\mathbf{x}.
\end{eqnarray} 

It is important to correctly preserve the motion boundaries in order to obtain an accurate optical flow. Motion boundaries are usually aligned with image boundaries, this aspect has motivated the use of edge detectors (\eg~\cite{dollar2013structured,leordeanu2012efficient}) in previous optical flow methods \cite{leordeanu2013locally,menze2015discrete,EpicFlow}.
Instead, we opt for using the  Non-Local Total Variation ($NLTV$) regularizer. $NLTV$ was  used for optical flow estimation in \cite{werlberger2010motion}, where the authors show its ability to better recover motion boundaries, in particular in low-textured areas, occluded regions, and in small objects (when used in a coarse-to-fine scheme). In our case, a non-local regularizer which better captures motion discontinuities is also very useful in the local minimization step where the initial correspondences are densified. 
We apply the $NLTV$ regularizer to each flow channel independently:
\begin{eqnarray} \label{eq:ER1}
E^1_R(\mathbf{u})  = \int_\Omega \int_\Omega  \omega(\mathbf{x},\mathbf{y}) \left(|u_1(\mathbf{x}) - u_1(\mathbf{y})|+|u_2(\mathbf{x}) - u_2(\mathbf{y})|\right) d\mathbf{y} d\mathbf{x},
\end{eqnarray}
where 
$ \omega(\x,\y) = \frac{1}{\mathcal{W}(\x)} e^{\frac{-\Delta_c(\x,\y)}{\sigma_c}} e^{\frac{-\Delta_s(\x,\y)}{\sigma_s}},$
$\Delta_c(\x,\y)$ denotes the Euclidean distance between the color values at $\x$ and $\y$ in the Lab space, $\Delta_s(\x,\y)$ is the Euclidean distance between points $\x$ and $\y$, $\mathcal{W}(\x)=\int_\Omega \omega(\x,\y) d\y $ is  a normalizing constant, and $\sigma_c$, $\sigma_s>0$ are constant parameters 
(set to $\sigma_c=2$ and $\sigma_s=2$).

As before, with the purpose of testing the proposed minimization procedure against different energies, we consider other regularization terms such as the classical 
coupled Total Variation (called $TV_{\ell_2}(u)$ in~\cite{strekalovskiy-et-al-siims14}), namely,
\begin{eqnarray} \label{eq:ER_TVcoupled}
E^2_R(\u)  =  \int_\Omega \|\nabla \u\|_2 d\x = \int_\Omega  \sqrt{|\nabla u_1(\x)|^2 + |\nabla u_2(\x)|^2}  d\x,
\end{eqnarray}

Appendix \ref{appendix:optalgos}  details the optimization algorithms for all the energies. Let us now just say that the data and the regularization terms are decoupled and standard methods such as primal-dual, thresholding or median schemes are used.
\subsection{Different possibilities for the initial set of seeds}
\label{ssec:ini_seeds}
\begin{figure*}[htbp]
\centering
  \subfloat[First frame]{
    \includegraphics[width=0.3\textwidth]{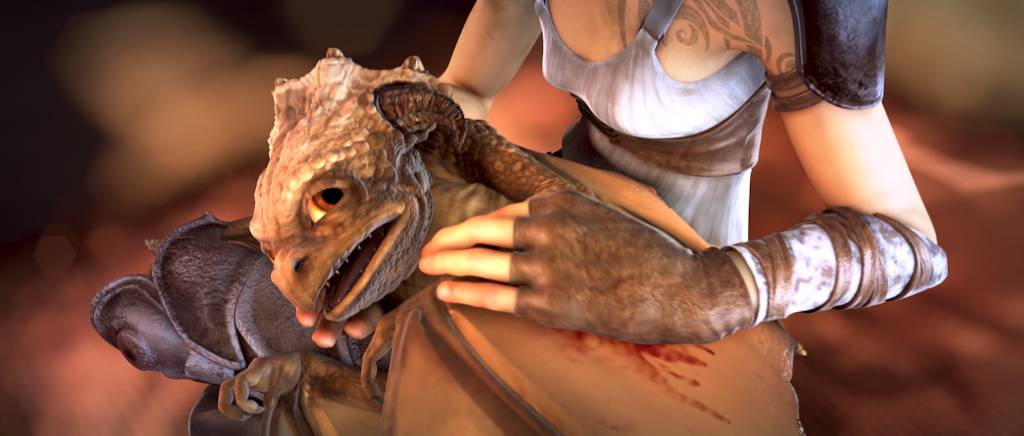}
    \label{fig:b1_F_35}}
  \subfloat[Second frame]{
    \includegraphics[width=0.3\textwidth]{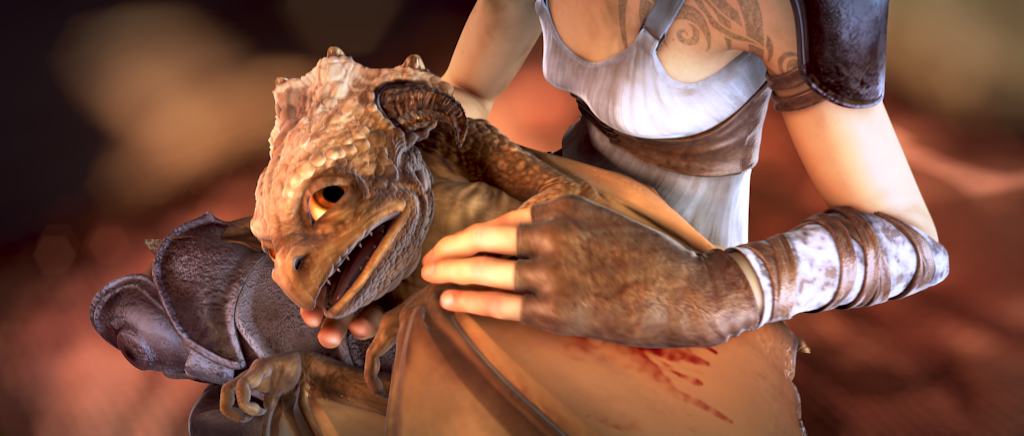}
    \label{fig:b1_F_36}}
  \subfloat[Ground truth]{
    \includegraphics[width=0.3\textwidth]{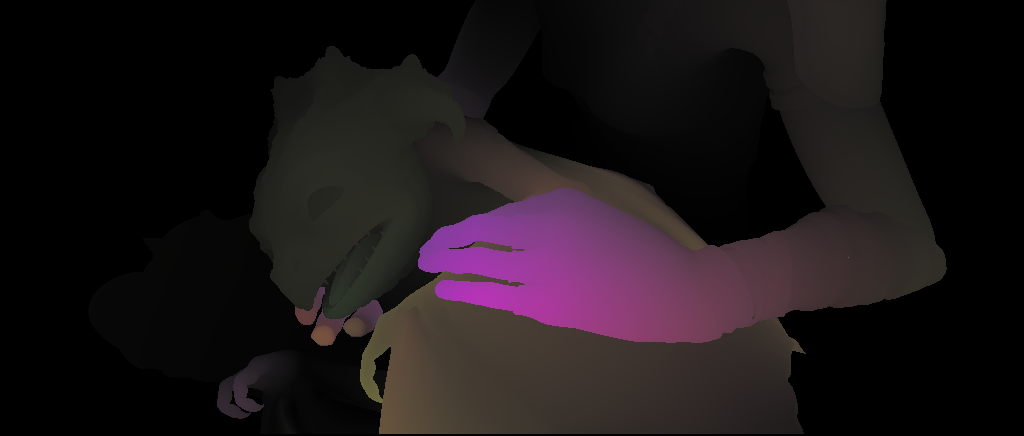}
    \label{fig:b1_F_35_GT}}
  \\
  \subfloat[SIFT seeds]{
    \includegraphics[width=0.3\textwidth]{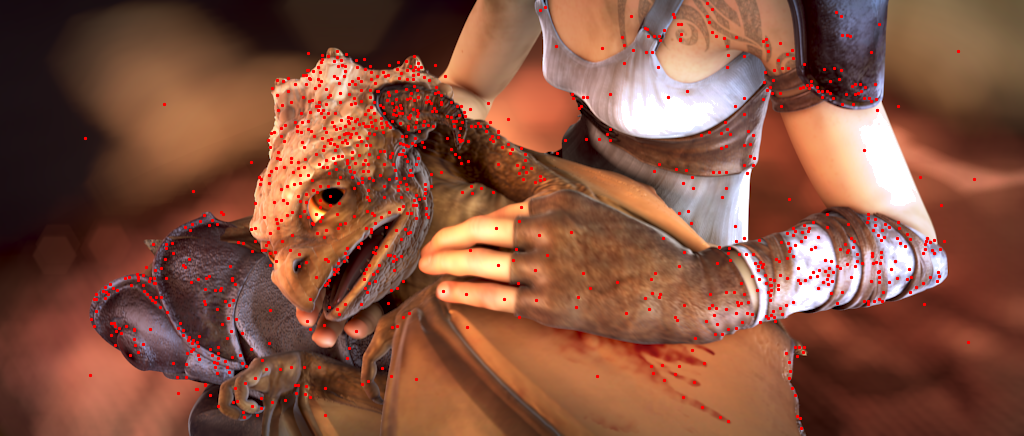}
  \label{fig:b1_F_35_sift_mt}}
  \subfloat[Local]{
    \includegraphics[width=0.3\textwidth]{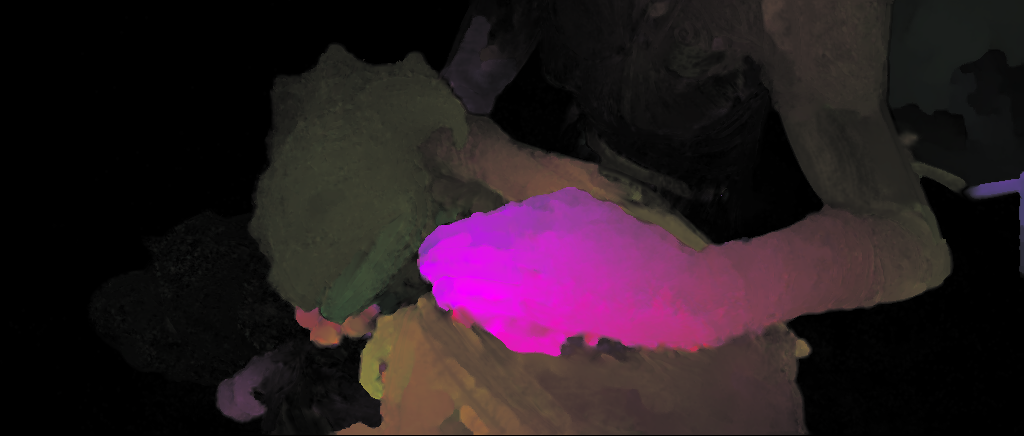}
  \label{fig:b1_F_35_sift_rg}}
  \subfloat[Global]{
    \includegraphics[width=0.3\textwidth]{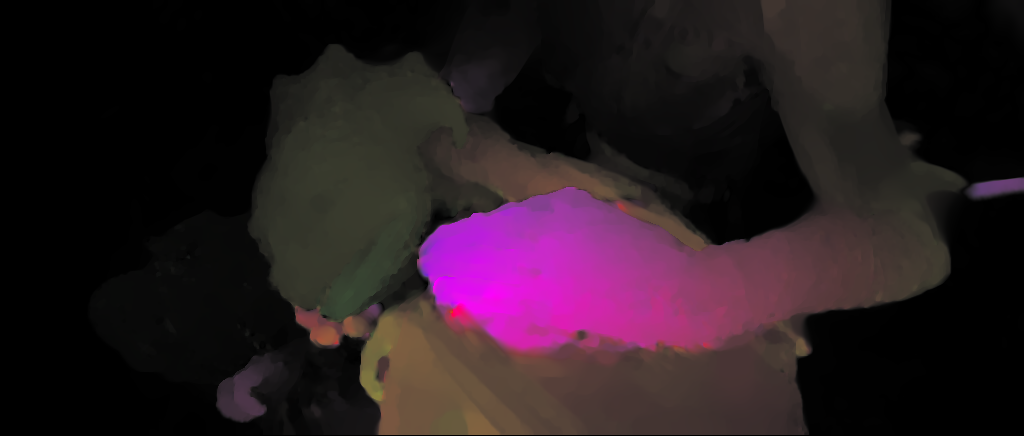}
  \label{fig:b1_F_35_sift_var}}
  \\
  \subfloat[DeepMatching seeds]{
    \includegraphics[width=0.3\textwidth]{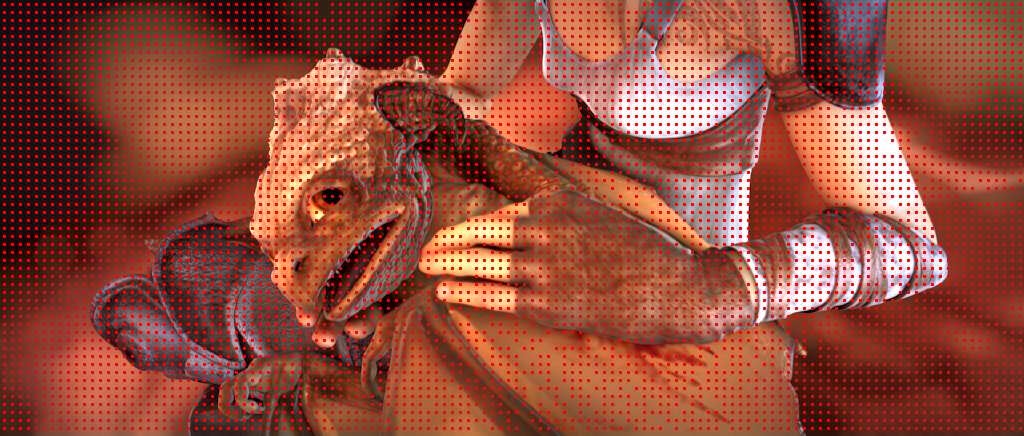}
   \label{fig:b1_F_35_deep_mt}}
  \subfloat[Local]{
    \includegraphics[width=0.3\textwidth]{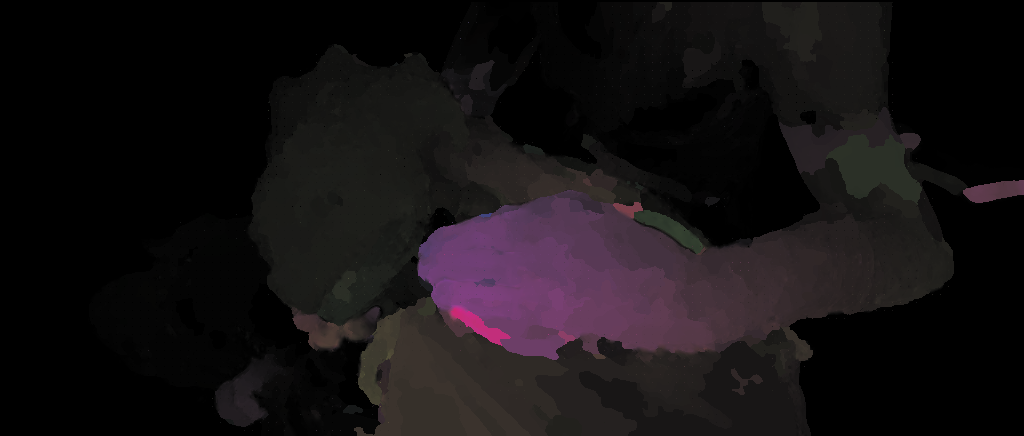}
  \label{fig:b1_F_35_deep_rg}}
  \subfloat[Global]{
    \includegraphics[width=0.3\textwidth]{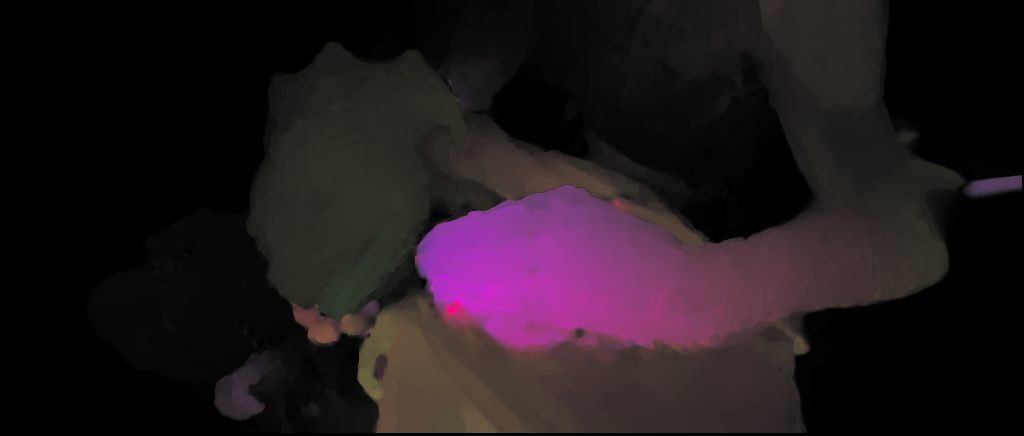}
  \label{fig:b1_F_35_deep_var}}
   \caption{Sequences with enough initial seeds at both trackers to recover a correct dense flow. The energy functional used is the classical $TV_{\ell_2}$-$L1$. }
   \label{fig:mpi_ini_easy_seeds}
\end{figure*}

\begin{figure*}[htbp]
  \centering
  \subfloat[First frame]{
    \includegraphics[width=0.3\textwidth]{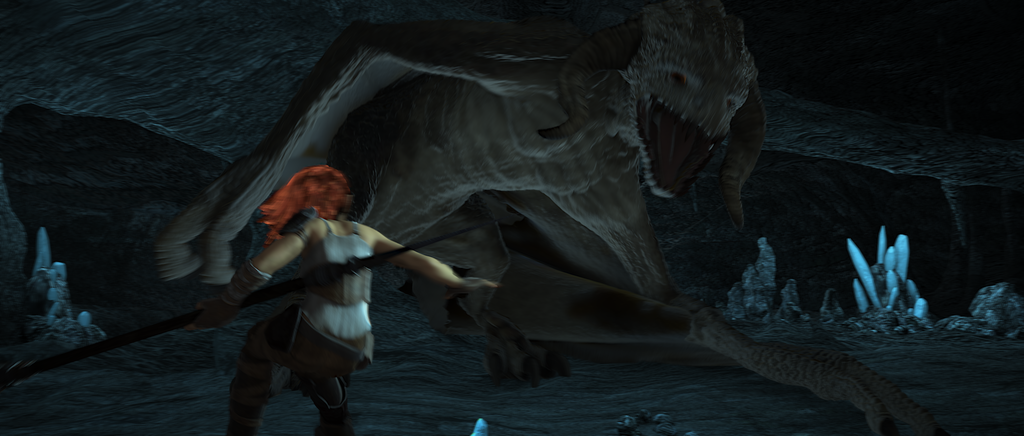}
  \label{fig:cave4_F_41}}
  \subfloat[Second frame]{
    \includegraphics[width=0.3\textwidth]{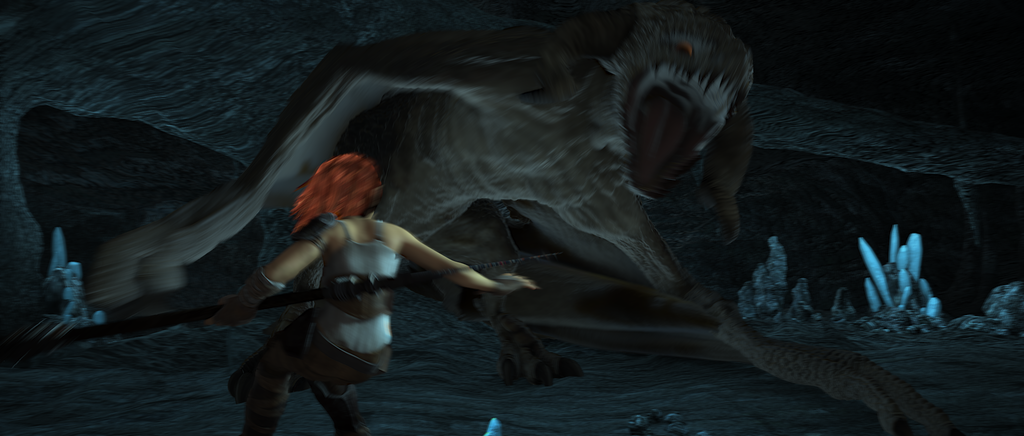}
  \label{fig:cave4_F_42}}
  \subfloat[Ground truth]{
    \includegraphics[width=0.3\textwidth]{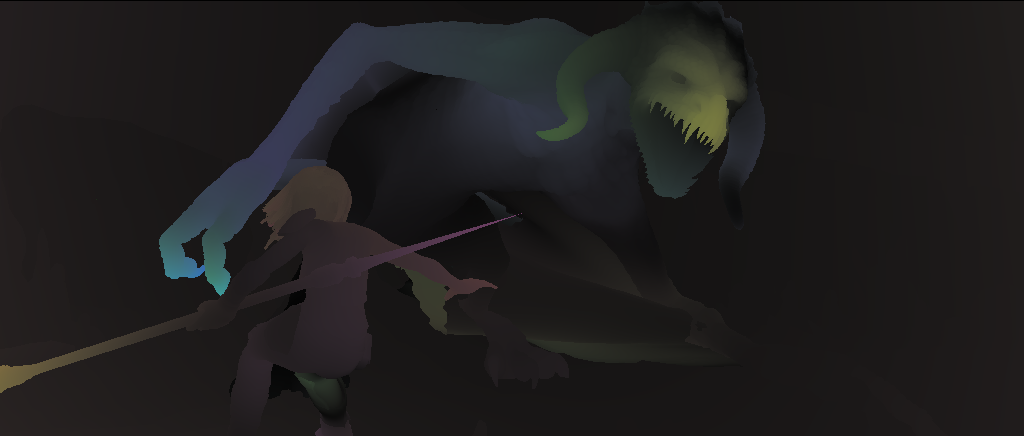}
  \label{fig:cave4_F_41_gt}}
  \\
  \subfloat[SIFT seeds]{
    \includegraphics[width=0.3\textwidth]{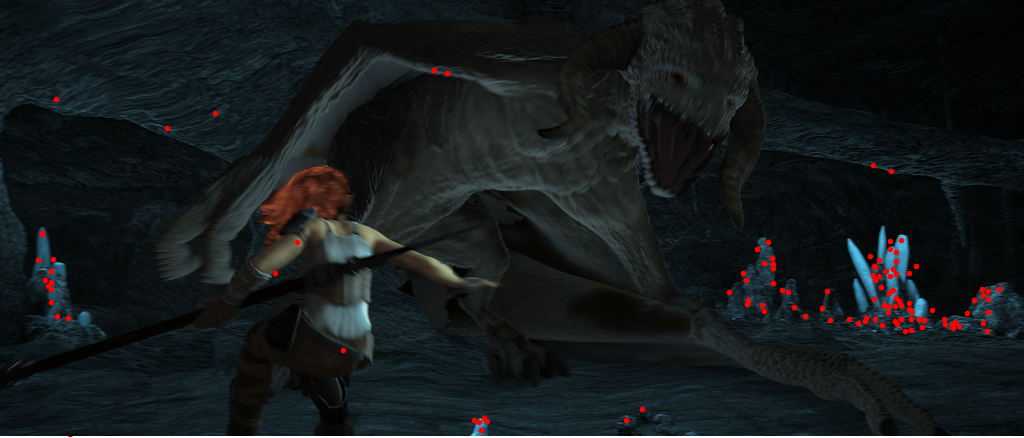}
  \label{fig:cave4_F_41_sift_mt}}
  \subfloat[Local]{
    \includegraphics[width=0.3\textwidth]{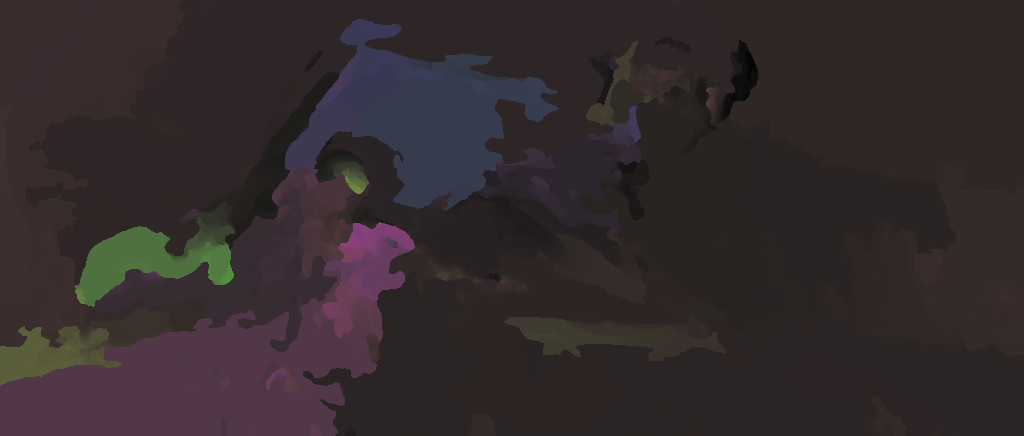}
  \label{fig:cave4_F_41_sift_rg}}
  \subfloat[Global]{
    \includegraphics[width=0.3\textwidth]{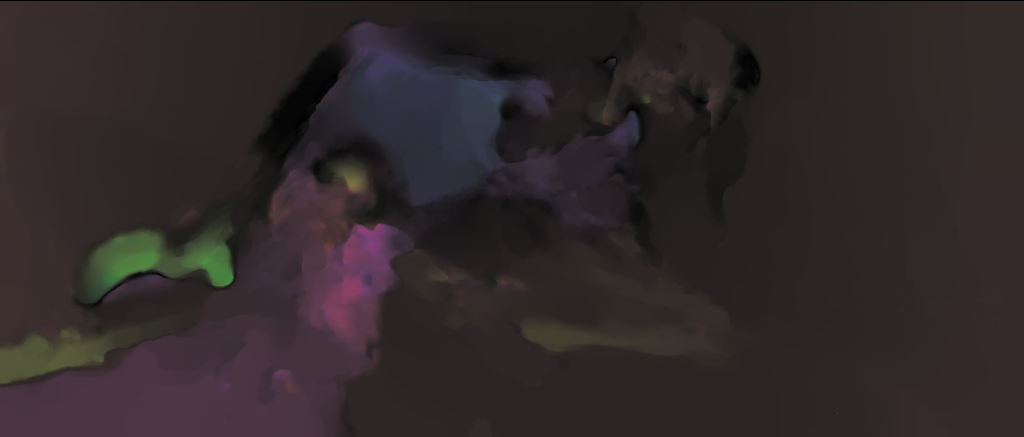}
  \label{fig:cave4_F_41_sift_var}}
  \\
  \subfloat[DeepMatching seeds]{
    \includegraphics[width=0.3\textwidth]{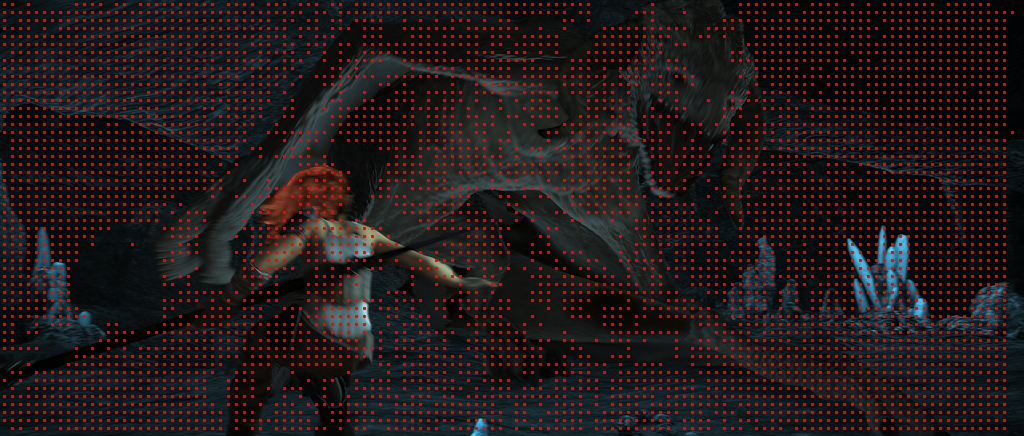}
  \label{fig:cave4_F_41_deep_mt}}
  \subfloat[Local]{
    \includegraphics[width=0.3\textwidth]{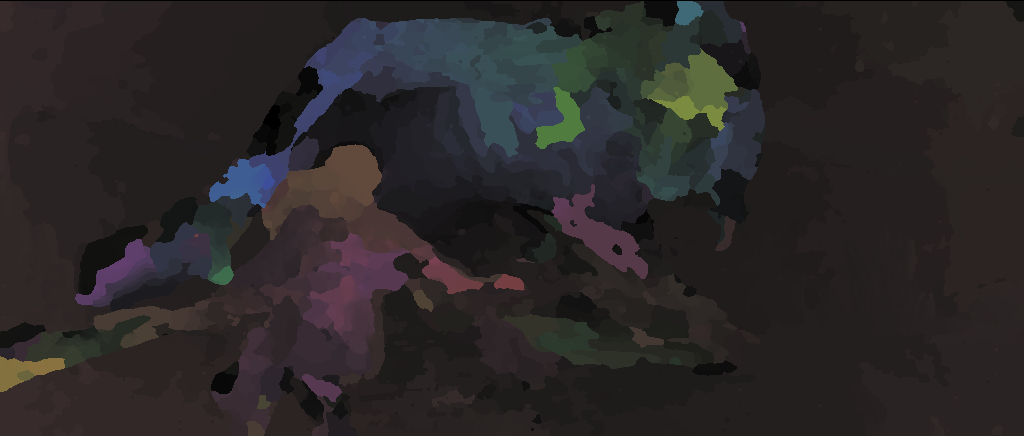}
  \label{fig:cave4_F_41_deep_rg}}
  \subfloat[Global]{
    \includegraphics[width=0.3\textwidth]{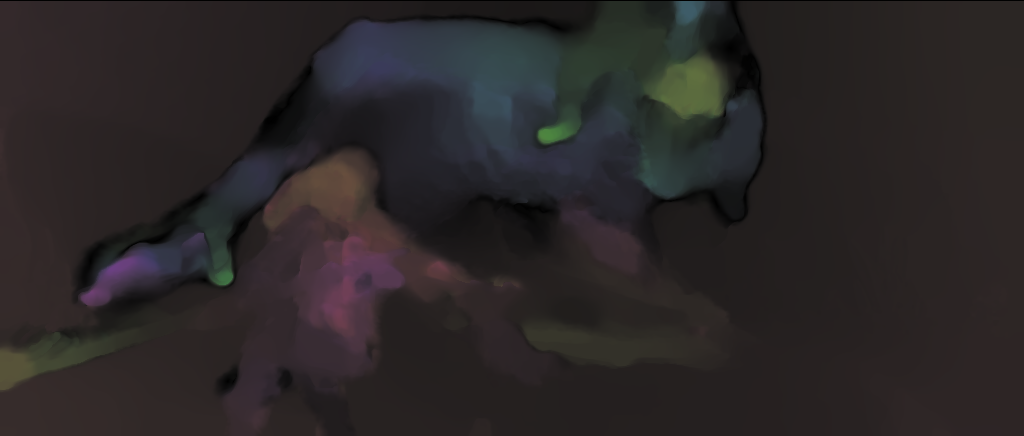}
  \label{fig:cave4_F_41_deep_var}}
  \caption{Sequence where only the second tracker (Deep Matching \cite{Weinzaepfel2013})  gets enough seeds to estimate a correct flow. The energy functional used is the classical $TV_{\ell_2}$-$L1$.}\label{fig:mpi_ini_hard_seeds}
\end{figure*}

In this work, the initial seeds are computed from one of the sparse matchers in the literature. 
There are several methods that provide a set of sparse matches between two different images containing common objects, representing two views of a scene or, as in our case, two frames of a video. Some of them are based in the estimation of distinctive point location and matching~\cite{SIFT,Mikolajczyk2005}. Being based on local properties, they can be used to estimate arbitrarily large displacements. In our experiments, we will use SIFT~\cite{SIFT} or Deep Matching~\cite{Weinzaepfel2013}.
Although SIFT is very effective, it has some disadvantages when dealing with small textureless objects or with non rigid deformations. The recent Deep Matching algorithm handles these problems and generates a more dense set of matchings. 
We have compared these two different methods, SIFT and Deep Matching, to compute the initial set of matches (the seeds). The first column of Figures~\ref{fig:mpi_ini_easy_seeds} and \ref{fig:mpi_ini_hard_seeds} show some examples of the set of seeds obtained by the SIFT and Deep Matching algorithms.
We use the SIFT implementation of \cite{Sift_Ipol}, with its default parameters. Also, we use the Deep Matching implementation with the default parameters presented in \cite{Weinzaepfel2013}.

With the aim of avoiding outliers in homogeneous areas, we initially remove the seeds having a low local saliency, which is determined by the minimum eigenvalue of the autocorrelation matrix computed locally. If the saliency is below a threshold (set to $0.045$ in our experiments), these seeds are removed. This pruning is also used by other authors (see, \eg, \cite{EpicFlow}).

In any case, we claim that our optical flow strategy is able to recover the dense motion regardless of the number of initial seeds; the only condition is to have at least one correct seed in every region $\Omega_{i}\subset\Omega$ where the motion is  smooth. As a proof of concept, in Fig.~\ref{fig:sintel_claim_alley1} we use a single seed per $\Omega_{i}$, which has been extracted from the ground truth flow. The estimated optical flow compares favourably to the ground truth even in these cases where the initial set of matches is extremely sparse. In contrast, other sparse-to-dense methods like \cite{EpicFlow} that produce state-of-the-art results are not capable to estimate the optical flow in this challenging situation. 
Another example is shown in Fig. \ref{fig:mpi_ini_easy_seeds} where similar optical flow results are obtained independently of the cardinality of the set of seeds, which have been computed using either the SIFT (Fig. \ref{fig:mpi_ini_easy_seeds}, second row) or DeepMatching \cite{Weinzaepfel2013}  (Fig. \ref{fig:mpi_ini_easy_seeds}, third row) algorithms. Let us notice that DeepMatching produces in general more matches than SIFT. 
On the other hand, Fig.~\ref{fig:mpi_ini_hard_seeds}  shows a situation where the lack of seeds in some regions $\Omega_{i}$ (second row, seeds provided by SIFT matches) produces an incorrect flow estimation in these areas. A much better estimation is obtained when there is at least one seed in these regions, as it happens when we use as seeds the matches provided by DeepMatching (Fig.~\ref{fig:mpi_ini_hard_seeds}, third row).

\subsection{Experimental results}
\label{sec:experiments}

The proposed method has been tested on several publicly available databases: Middlebury \cite{Middlebury}, MPI-Sintel \cite{Sintel}, KITTI 2012 \cite{Kitti}, and KITTI 2015 \cite{Menze2015CVPR}, as well as on proof of concept examples, chosen in order to obtain a better understanding of the fundamental parts of the proposed strategy. 
Let us remark that all results have been obtained by using the grayscale versions of the original color frames. The  color version is only used to compute the seeds in case they come from the Deep Matching algorithm and to compute the local regularization weights in case of the Non-Local Total Variation as regularization term. All the results in this section have been obtained with the iterated faldoi minimization strategy presented in Sect.~\ref{ssec:iteratedfaldoi}.
We have fixed the parameters for all the experiments, more details are given in Appendix \ref{appendix:impldetails}. 

\begin{figure*}[htbp]
\centering
\subfloat[First frame]{
    \includegraphics[width=0.3\textwidth]{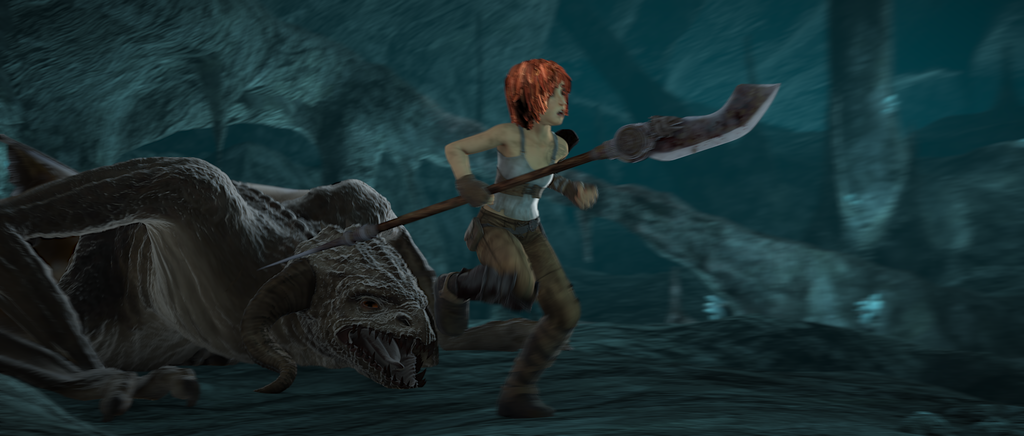}
  }
  \subfloat[Second frame]{
    \includegraphics[width=0.3\textwidth]{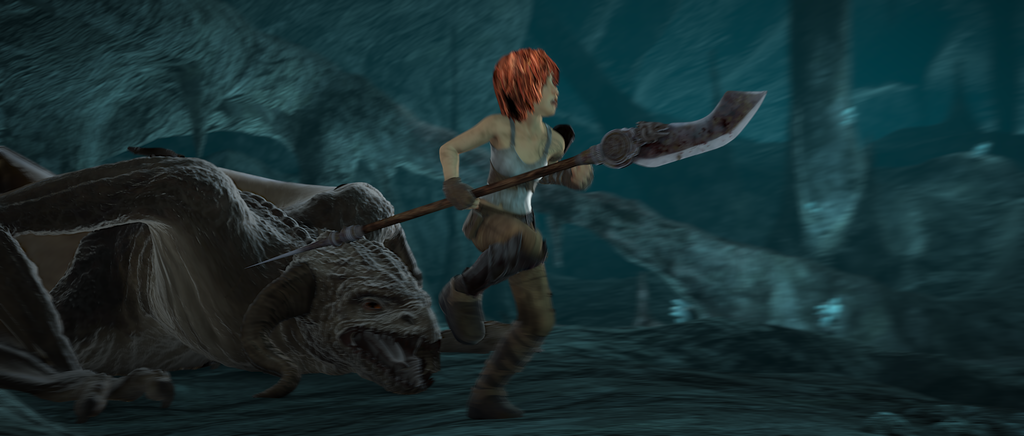}
  }
    \subfloat[Ground truth]{
    \includegraphics[width=0.3\textwidth]{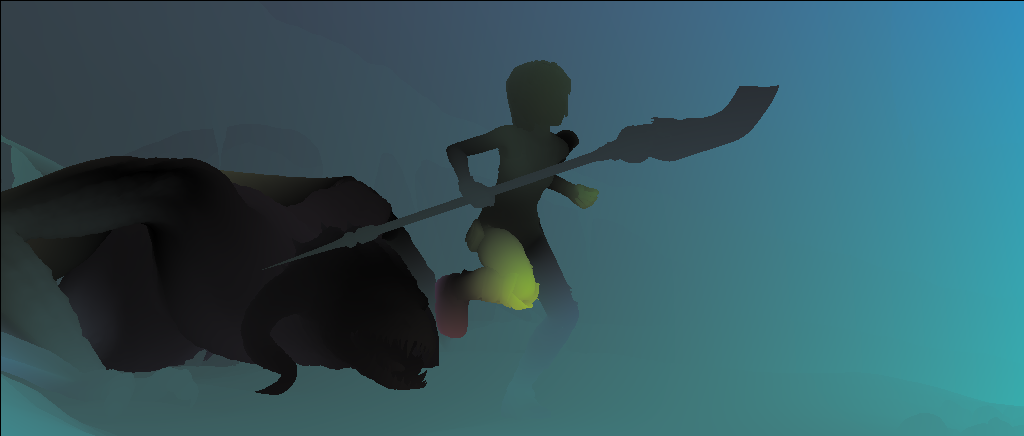}
    }
  \\
\subfloat[Occlusion map]{
    \includegraphics[width=0.3\textwidth]{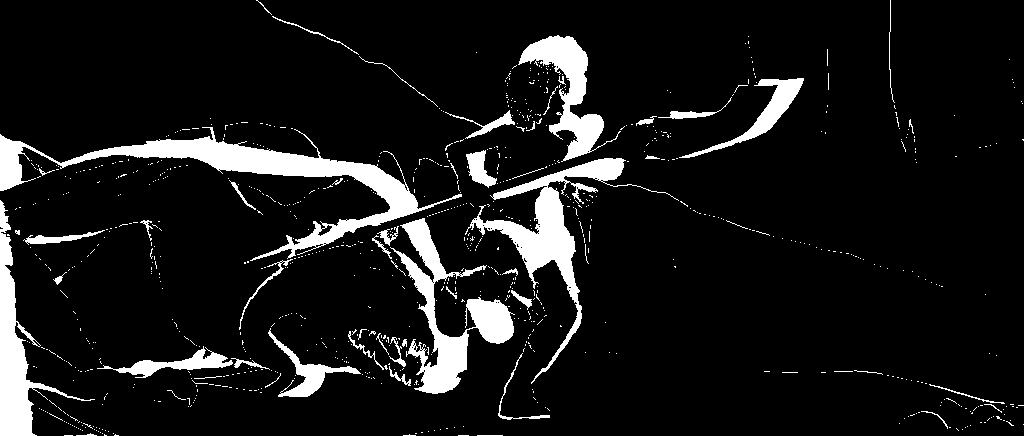}
    }
  \subfloat[Multi-scale $TV_{\ell_2}$-$L1$]{
    \includegraphics[width=0.3\textwidth]{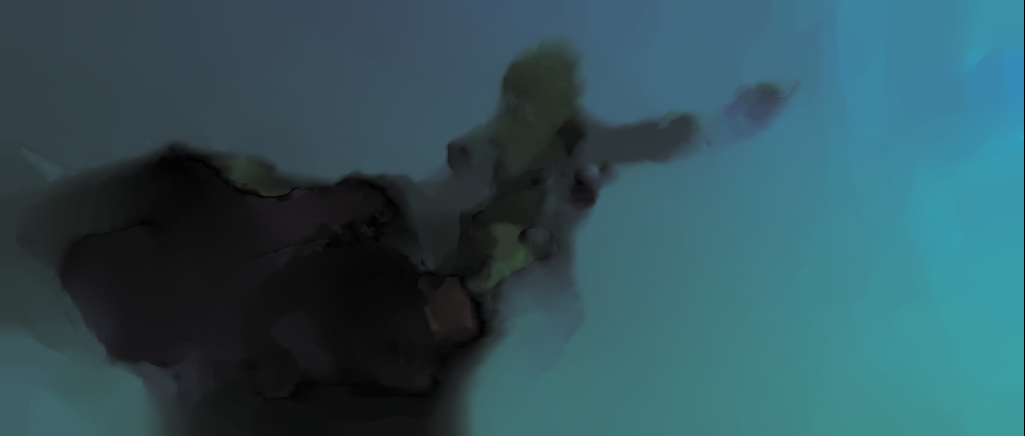}
    }
  \subfloat[Multi-scale $NLTV$-$CSAD$]{
    \includegraphics[width=0.3\textwidth]{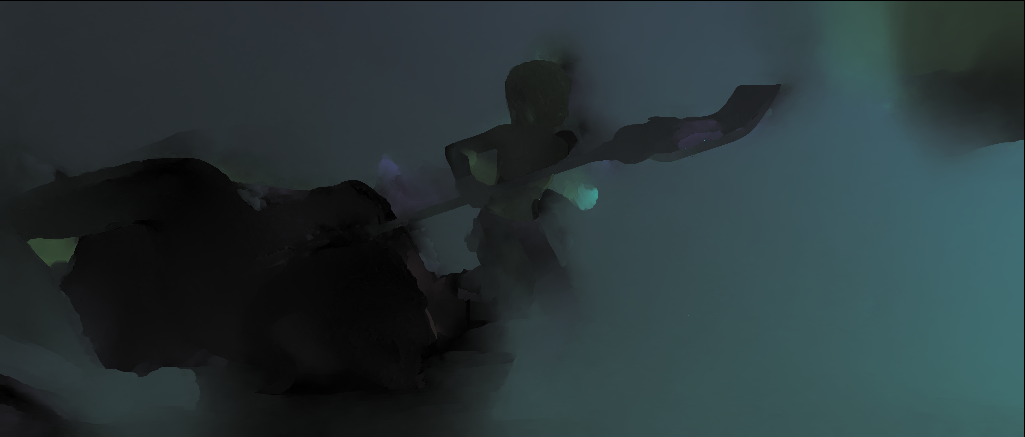}
    }
    \\
\subfloat[Faldoi:$TV_{\ell_2}$-$CSAD$]{
    \includegraphics[width=0.3\textwidth]{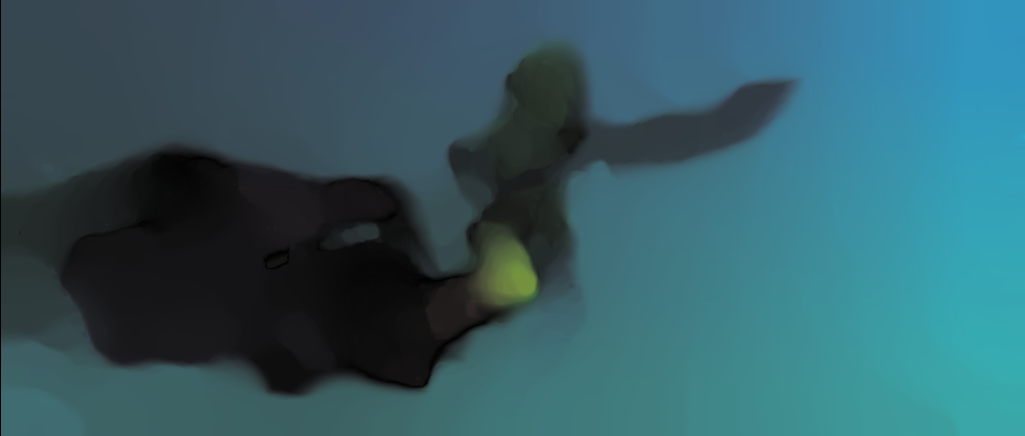}
    }
   \subfloat[Faldoi:$TV_{\ell_2}$-$L1$]{
    \includegraphics[width=0.3\textwidth]{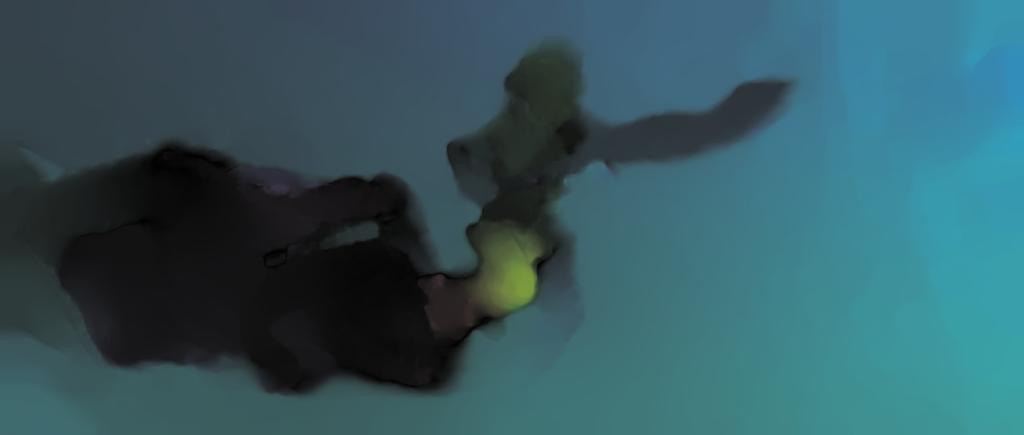}
    }
  \subfloat[Faldoi:$NLTV$-$CSAD$]{
    \includegraphics[width=0.3\textwidth]{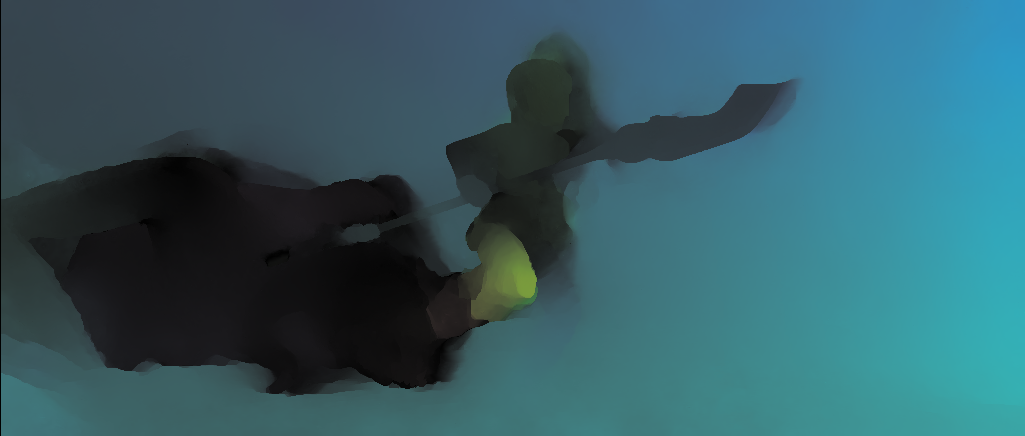}
    }
  \caption{Comparison of our strategy against the multi-scale approach using different functionals. The initial seeds have been computed using DeepMatching.}\label{fig:functional_comparison_c2}
\end{figure*}

First, we present a comparison of our strategy against the multi-scale approach using different functionals. 
In order to assure that the proposed framework is a real alternative to the coarse-to-fine warping strategy and, therefore, also valid for sequences that do not have large displacements of small objects, experiments on the Middlebury optical flow data set \cite{Middlebury} have been performed for both approaches. Table \ref{table:MIdd_EPE-AAE-common} shows how our framework achieves better results in all the samples in the dataset and for both energy functionals, $TV_{\ell_2}$-$L1$ and  $NLTV$-$CSAD$, even if the difference is slight is some cases. 
Some qualitative results are shown for images of the MPI Sintel database that contain large displacements in Figure ~\ref{fig:functional_comparison_c2} 
where the initial seeds have been computed using DeepMatching. 
The first row shows, from left to right, two consecutive frames and the optical flow ground truth (color coded). The image (d) of the second row shows the ground truth occlusion map (the occluded points are shown in white). The multi-scale results obtained using the $TV_{\ell_2}$-$L1$ and the $NLTV$-$CSAD$ energies are shown in (e) and (f), respectively, while the corresponding results obtained by our minimization  strategy (iterated faldoi) are shown in (h) and (i), respectively. Finally, image (g) displays the result of our minimization faldoi strategy applied to the $TV_{\ell_2}$-$CSAD$ energy functional. As it can be observed, the use of an advanced data term based on patches reduces the 
outlier area and better recovers the human shape. After adding the non-local regularizer the motion 
boundaries are more accurate and the outlier is  removed.

\begin{table*}
\begin{center}
\begin{tabular}{|lllllllll|c|}
\hline
Middlebury     & Dim.     & Hyd.     & Rub.     & Gro2     & Gro3     & Urb2     & Urb3     & Ven \\
\hline\hline
$TV_{\ell_2}$-$L1$, multi-scale & $0.1537$ & $0.2286$ & $0.1916$ & $0.1496$ & $0.6808$ & $0.3709$ & $0.6034$ & $0.3563$\\
$TV_{\ell_2}$-$L1$, our min. & $0.1243$ & $0.2078$ & $0.1876$ & $0.1397$ & $0.5945$ & $0.3599$ & $0.4354$ & $0.3109$\\
$NLTV$-$CSAD$, multi-scale & $0.1286$ & $0.2003$ & $0.1509$ & $0.1405$ & $0.5652$ & $0.4315$ & $0.5763$ & $0.3647$\\
$NLTV$-$CSAD$, our min.& $0.1075$ & $0.1984$ & $0.1477$ & $0.1389$ & $0.5414$ & $0.4028$ & $0.5521$ & $0.2961$\\
\hline
\end{tabular}
\end{center}
\caption{Error measures in the Middlebury dataset with public ground truth. The first two lines correspond to the $TV_{\ell_2}$-$L1$ energy functional while the last two lines are based on the $NLTV$-$CSAD$ functional. The results obtained minimizing with the classical coarse-to-fine scheme are shown in the first and third line and the ones obtained minimizing with our proposal are shown in the second and fourth line.}\label{table:MIdd_EPE-AAE-common}
\end{table*}

\begin{figure*}[htbp]
\centering
  \subfloat[$TV_{\ell_2}$-$L1$]{
    \includegraphics[width=0.50\textwidth]{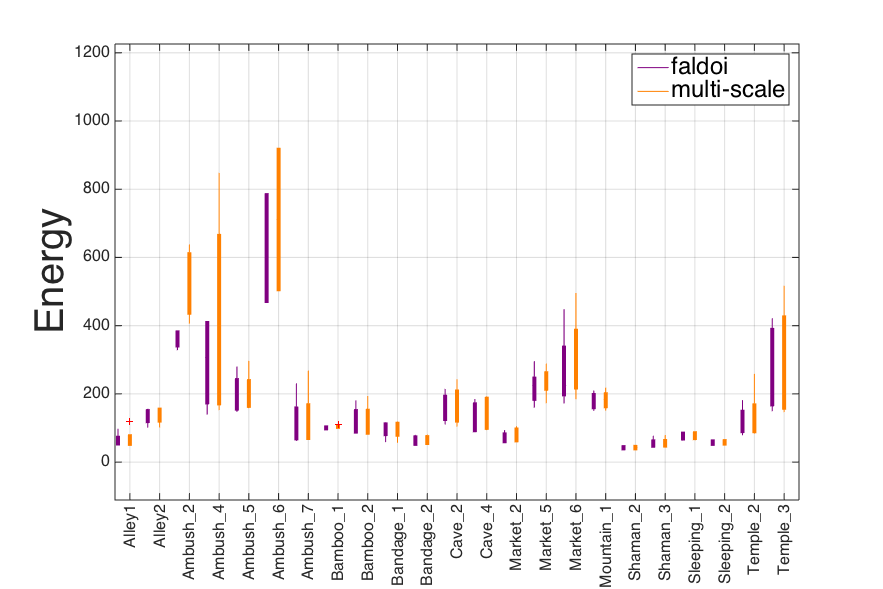}
  }
  \subfloat[$NLTV$-$CSAD$]{
    \includegraphics[width=0.50\textwidth]{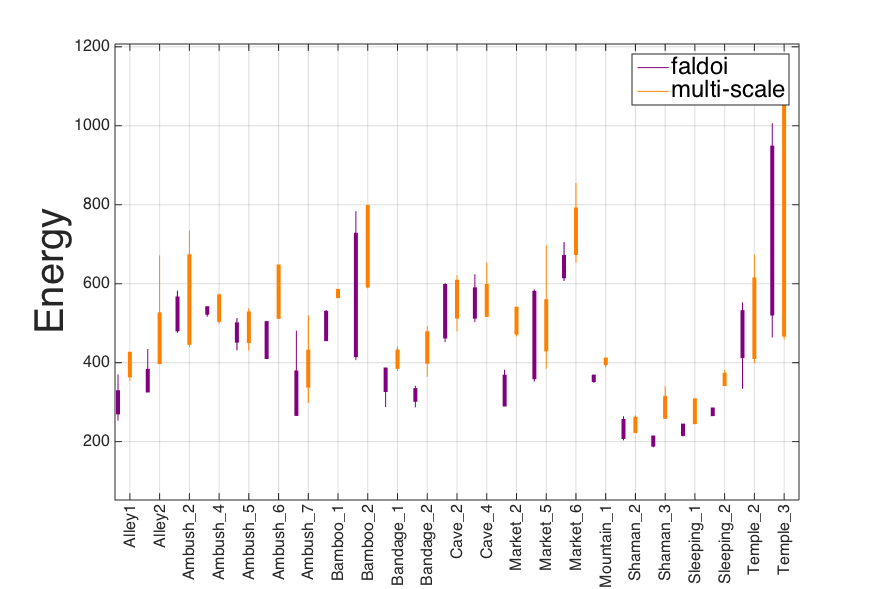}
  }
  \caption{Energy results for $TV_{\ell_2}$-$L1$ and $NLTV$-$CSAD$ using the multi-scale approach and our proposal on a randomly selected subset  of the MPI-Sintel \textit{Final} training set.
  It is the standard boxplot representation. Red dots are outliers, each box shows the first, second and third quartiles and the whiskers length corresponds to 1.5 interquartile range (IQR).
  }\label{fig:energy_table}
\end{figure*}

Table~\ref{table:set1} provides a quantitative comparison of different estimations of the optical flow obtained with different energy functionals ($TV_{\ell_2}$-$L1$ and $NLTV$-$CSAD$) and different set of seeds (SIFT seeds and Deep Matching seeds). We use a subset of 10\% of frames randomly selected from the final training set of the MPI Sintel database. Let us notice that the energy based on the non-local Total Variation and a smooth approximation of the Census transform ($CSAD$) is the one which achieves the best results, both quantitative and qualitatively. On the other hand, Figure~\ref{fig:energy_table} shows the energy values using both approaches, multi-scale and our proposal (with Deep Matching seeds) over the same functionals. Notice how in general our method gives a lower or similar energy than the multi-scale method.

\begin{table*}
\begin{center}
\begin{tabular}{|lllllll|}
\hline
Functional, minimization & EPE all & EPE mat. & EPE unmat. & s0-10 & s10-40 & s40+ \\
\hline\hline
$TV_{\ell_2}$-$L1$, multi-scale &  6.6453 &  5.1181  & 15.8828 &  2.9261 &  8.0172 &  45.3955  \\
$NLTV$-$CSAD$, multi-scale & 8.4243 &  7.0860 &  16.8037 &  2.8129 &  9.3389 &  50.4899\\
$TV_{\ell_2}$-$L1$, our min.~(SIFT)   &    6.7307   & 5.4784  & 14.8439 &    3.2928 & 8.3336 &  47.2940\\
$NLTV$-$CSAD$, our min.~(SIFT)  &  7.3337  &  6.1058  & 14.7972  &   4.4182  &  10.1687 &  49.8888\\
$TV_{\ell_2}$-$L1$, our min.~(Deep)  & 4.7688  &  3.5043 &  13.5503 &    2.3861 &    6.4305 &  32.3350\\
$TV_{\ell_2}$-$CSAD$, our min.~(Deep)  &    4.0158  &  2.7901 &  12.5350 &  2.0757 &  5.4551 &  31.5814\\
$NLTV$-$CSAD$, our min.~ (Deep)  &     4.0060  &  2.7845 &  12.2698 & 2.3115 &  5.2989 &  30.8537  \\
\hline
\end{tabular}
\end{center}
\caption{Results for several methods (coarse-to-fine and our proposal), several energies ($TV_{\ell_2}$-$L1$ and $NLTV$-$CSAD$) and several initial sets of seeds (SIFT seeds and Deep Matching seeds) on a randomly selected subset  of the MPI-Sintel \textit{Final} training set.}
\label{table:set1}
\end{table*}

\begin{figure}
\begin{tabular}{cc}
  \includegraphics[width=0.45\linewidth]{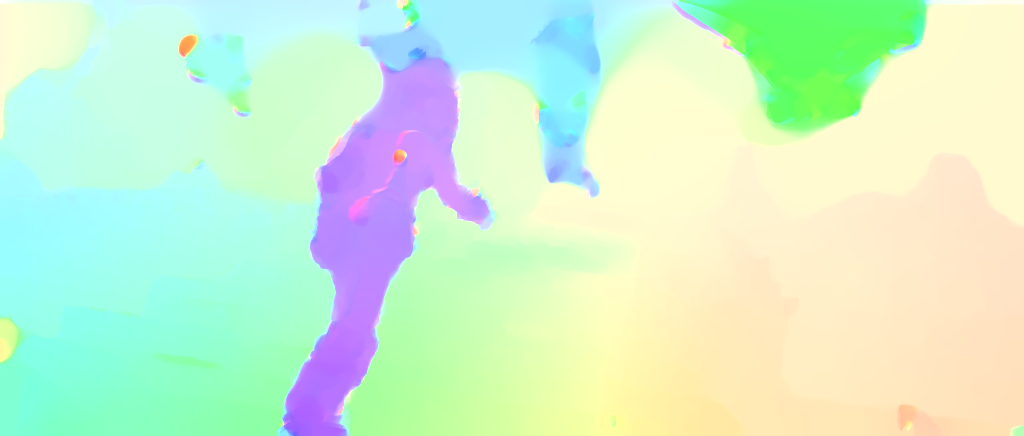}&
 \includegraphics[width=0.45\linewidth]{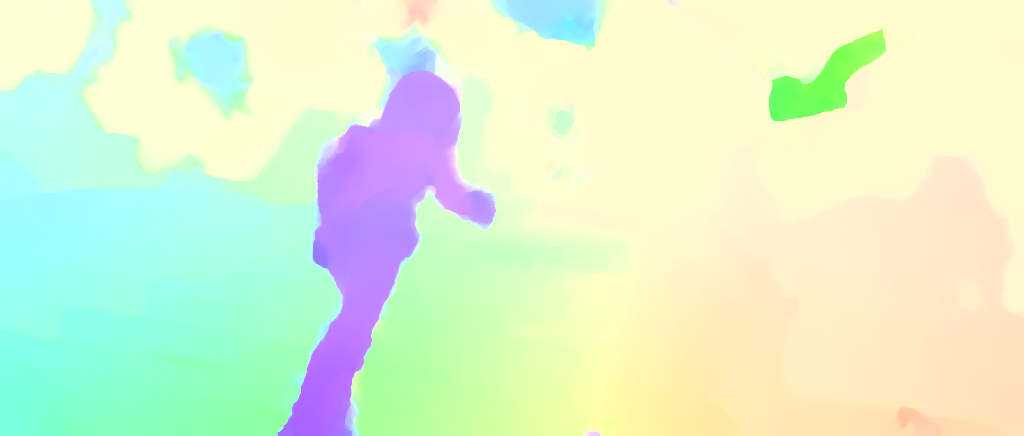}\\
     LDOF \cite{Brox_Malik_LD}, $EPE=1.460$ & Discrete Flow \cite{menze2015discrete}, $EPE=0.788$ \\
     \includegraphics[width=0.45\linewidth]{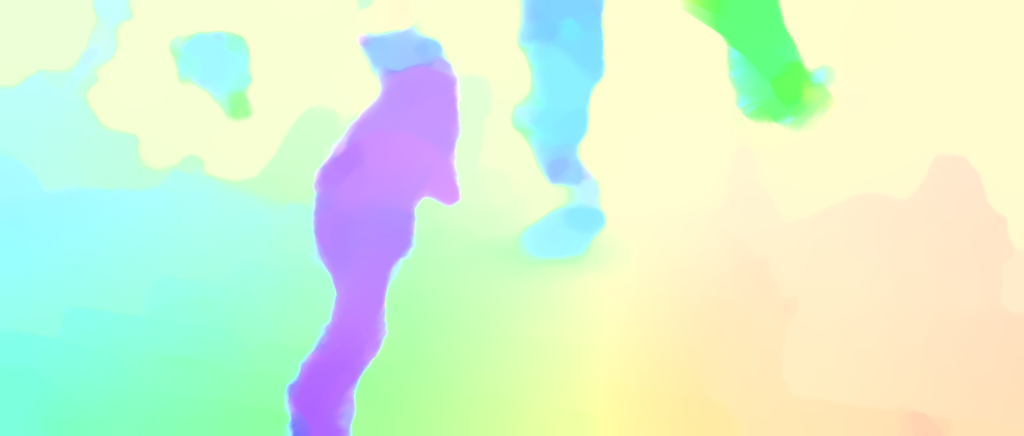}&
    \includegraphics[width=0.45\linewidth]{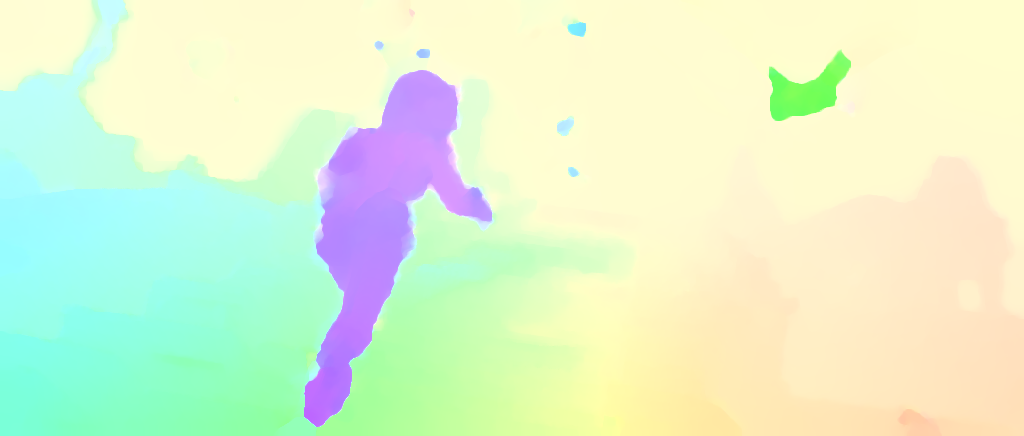}\\
    DeepFlow \cite{Weinzaepfel2013}, $EPE=1.230$ & EpicFlow \cite{EpicFlow}, $EPE=0.818$ \\
   \includegraphics[width=0.45\linewidth]{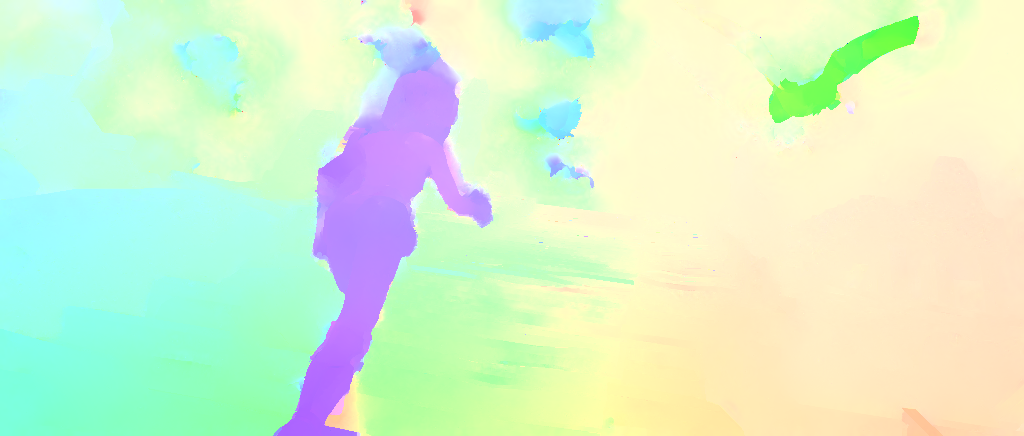}&
   \includegraphics[width=0.45\linewidth]{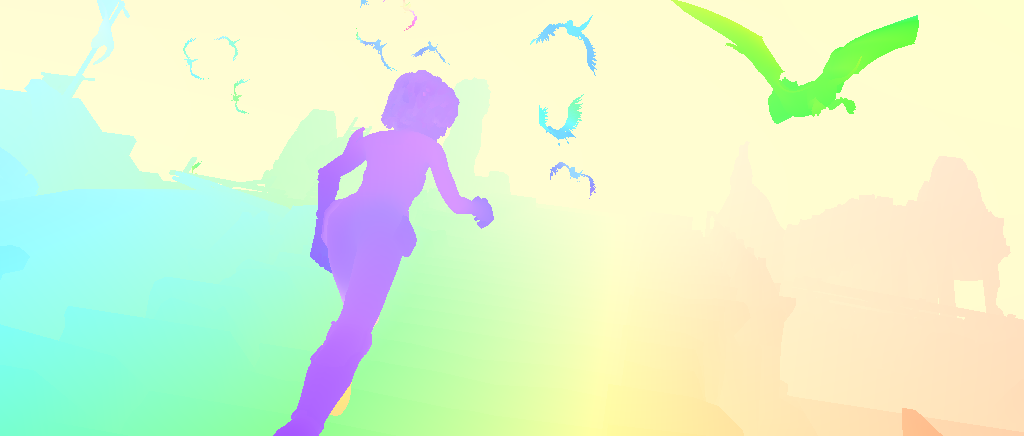}\\
   Our proposal, $EPE=1.095$ & Ground truth \\
 \end{tabular}
   \caption{Qualitative and quantitative comparison of different optical flow methods in a frame of the MPI-Sintel database (clean test). In our proposal we use the $NLTV$-$CSAD$ energy with Deep Matching seeds.}
\label{fig:comparison}
\end{figure}

Table~\ref{table:SintelComparison} and Figure~\ref{fig:comparison} show a quantitative and a qualitative comparison, respectively, among our method and several state-of-the-art methods, 
including methods based on the combination of sparse matches and variational techniques \cite{Brox_Malik_LD,Weinzaepfel2013} via an extra coupling term and that can also be adapted to any energy. Our approach outperforms LDOF \cite{Brox_Malik_LD} in both Clean and Final versions of the MPI-Sintel dataset. Compared to DeepFlow \cite{Weinzaepfel2013} we get better results in the Clean version and in the Final version we get an EPE which is only a tenth worse, thus we can say that the results are comparable in the Sintel database. 
Our approach gives better results than S2D-Matching \cite{leordeanu2013locally}, which is another sparse-to-dense technique that uses edge information and occlusion estimation in the densification process.
The images in Fig.~\ref{fig:comparison} have been taken from the MPI-Sintel webpage. 
Compared to \cite{Brox_Malik_LD,menze2015discrete,Weinzaepfel2013} and EpicFlow \cite{EpicFlow} (also a sparse-to-dense technique that achieves good positions in the Sintel table) our result better recovers the silhouettes of the girl (elbows, armpit, right hip and lower part of the legs) and the birds without using precomputed edge information as \cite{EpicFlow}, although it produces some halos around the birds and the top-left corner of the image. 

\begin{table*}
\begin{center}
\begin{tabular}{|lllllll|c|}
\hline
 & EPE all & EPE mat. & EPE unmat. & s0-10 & s10-40 & s40+ \\
\hline\hline
\noalign{\smallskip}
$Final$ \\
\hline
$FlowFields^{[4]}$ \cite{bailer2015flowICCV} & 5.810 & 2.621 & 31.799 & 1.157 & 3.739 & 33.890 \\
$DiscreteFlow^{[8]}$ \cite{menze2015discrete} & 6.077 & 2.937 & 31.685 & 1.074 & 3.832 & 36.339 \\
$EpicFlow^{[10]}$ \cite{EpicFlow}  & 6.285 & 3.060 & 32.564 & 1.135 & 3.727 & 38.021\\
$DeepFlow^{[16]}$ \cite{Weinzaepfel2013}  & 7.212 & 3.336 & 38.781 & 1.284 & 4.107 & 44.118 \\
$Our\; approach^{[21]}$ & 7.337 &	3.580 &	37.904	&1.487 &	4.355	&43.526	 \\
$S2D-Matching^{[27]}$ \cite{leordeanu2013locally}  & 7.872 & 3.918 & 40.093 & 1.172 & 4.695 & 48.782  \\
$LDOF^{[49]}$~\cite{Brox_Malik_LD}  & 9.116 & 5.037 & 42.344 & 1.485 & 4.839 & 57.296 \\
\hline
\hline\noalign{\smallskip}
$Clean$ \\
\hline
\noalign{\smallskip}
$DiscreteFlow^{[4]}$ \cite{menze2015discrete} & 3.567 & 1.108 & 23.626 & 0.703 & 2.277 & 20.906 \\
$FlowFields^{[5]}$ \cite{bailer2015flowICCV} & 3.748 & 1.056 & 25.700 & 0.546 & 2.110 & 23.602 \\
$EpicFlow^{[8]}$ \cite{EpicFlow}  & 4.115 & 1.360 & 26.595 & 0.712 & 2.117 & 25.859 \\
$Our\; approach^{[14]}$ & 4.927	& 1.542	& 32.535 &	1.047	&2.647	& 29.719	 \\
$DeepFlow^{[20]}$ \cite{Weinzaepfel2013}  & 5.377 & 1.771 & 34.751 & 0.960 & 2.730 & 33.701 \\
$S2D$-$Matching^{[38]}$ \cite{leordeanu2013locally}  & 6.510 & 2.792 & 36.785 & 0.622 & 3.012 & 44.187  \\
$LDOF^{[51]}$~\cite{Brox_Malik_LD}  & 7.563 & 3.432 & 41.170 & 0.936 & 2.908 & 51.696 \\
\hline
\end{tabular}
\end{center}
\caption{Results on MPI-Sintel test set (9th of Jun. of 2016). The first and second set of results correspond, respectively, to the \textit{Final} and \textit{Clean} frames. For our results we used the $NLTV$-$CSAD$ energy.}\label{table:SintelComparison}
\end{table*}

Table~\ref{table:K12Comparison} shows a quantitative comparison among our method and several methods on the KITTI 2012 database~\cite{Kitti}.
As expected, the performance of our approach is better  than LDOF~\cite{Brox_Malik_LD},  and comparable to CRTflow~\cite{DBLP:journals/ijcv/DemetzHW15}, EpicFlow~\cite{EpicFlow} and DeepFlow~\cite{Weinzaepfel2013}. KITTI 2015 \cite{Menze2015CVPR} can be used to evaluate optical flow and scene flow algorithms. Table~\ref{table:K15Comparison} only contains results for optical flow methods.

\begin{table*}
\begin{center}
\begin{tabular}{|lllll|c|}
\hline
 & EPE-noc & EPE-all & Out-Noc3 & Out-All 3 \\
\hline\hline
\noalign{\smallskip}
\hline
$FlowFields^{[22]}$ \cite{bailer2015flowICCV} & 1.4 & 3.5 & 5.77 &	14.01\\
$DiscreteFlow^{[27]}$ \cite{menze2015discrete}& 1.3 & 3.6  & 6.23 & 16.63 \\
$DeepFlow^{[31]}$ \cite{Weinzaepfel2013} & 1.5 & 5.8 & 7.22 & 17.79  \\
$EpicFlow^{[32]}$ \cite{EpicFlow}  & 1.5 & 3.8 & 7.88 & 17.08\\
$Our\; approach^{[35]}$ & 1.7 & 5 & 8.81 & 19.93\\
$CRTflow^{[37]}$ \cite{DBLP:journals/ijcv/DemetzHW15}  & 2.7 &  6.5   & 9.43 & 18.72\\
$LDOF^{[59]}$~\cite{Brox_Malik_LD} &  5.6 & 12.4  & 21.93 & 31.39\\
\hline
\end{tabular}
\end{center}
\caption{Results on Kitti 2012 test (14th of Sep. of 2016). Out-Noc3 (resp. Out-all3) refers to the percentage of pixels where the estimated optical flow presents an error above 3 pixels in non-occluded areas(resp. all pixels). EPE-noc is the EPE over non-occluede areas and EPE-all is over all the image. For our results we used the $NLTV$-$CSAD$ energy.}\label{table:K12Comparison}
\end{table*}

\begin{table*}
\begin{center}
\begin{tabular}{|llll|c|}
\hline
 & Fl-bg & Fl-bf & Fl-all \\
\hline\hline
\noalign{\smallskip}
\hline
$PatchBatch^[14]$ \cite{DBLP:journals/corr/GadotW15} & 19.98 & 30.24 & 21.69\\
$DiscreteFlow^{[17]}$ \cite{menze2015discrete}& 21.53 &	26.68 &	22.38 \\
$CPM-Flow^{[19]}$ \cite{Hu_2016_CVPR} & 22.32 &	27.79 &	23.23 \\
$EpicFlow^{[23]}$ \cite{EpicFlow}  & 25.81 & 33.56 &	27.10 \\
$Our\; approach^{[24]}$ & 27.08 & 31.51 &	27.81\\
$DeepFlow^{[26]}$ \cite{Weinzaepfel2013} & 27.96 & 35.28 & 29.18 \\
$HS^{[31]}$~\cite{sun2014quantitative} &  39.90 & 53.59 & 42.18\\
\hline
\end{tabular}
\end{center}
\caption{Results on Kitti 2015 test (14th of Sep. of 2016). Fl refers the percentage of optical flow outliers. bg (resp. fg) refers of percentage of outliers only over bakground regions (resp. foreground regions) and all means over all ground truth pixels. For our results we used the $NLTV$-$CSAD$ energy.}\label{table:K15Comparison}
\end{table*}

\begin{figure*}[htbp]
\centering
\subfloat[First frame]{
    \includegraphics[width=0.31\textwidth]{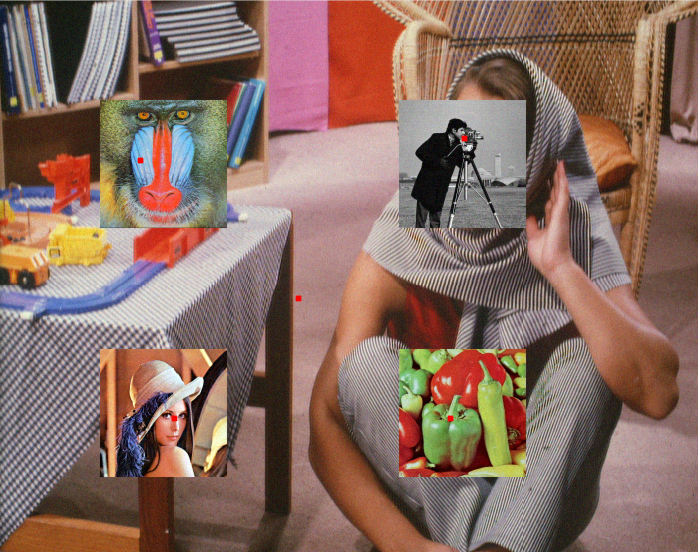}
  \label{fig:hm0_a}}
  \subfloat[Second frame]{
    \includegraphics[width=0.3\textwidth]{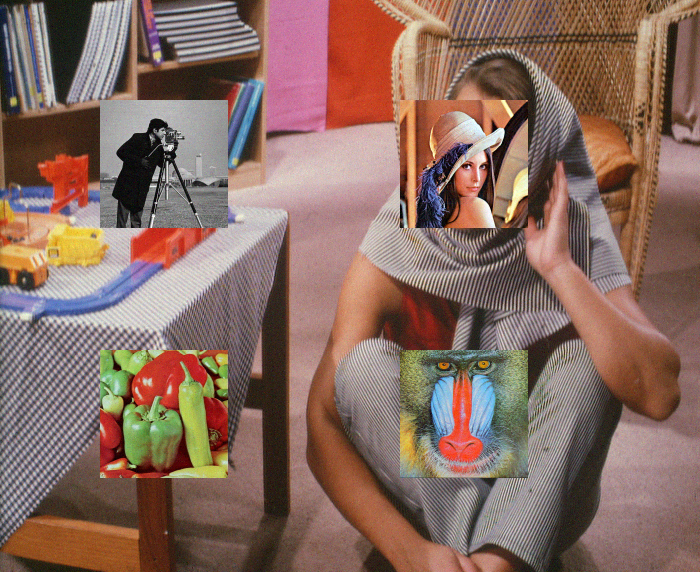}
  \label{fig:hm0_b}}
  \subfloat[Ground truth]{
    \includegraphics[width=0.3\textwidth]{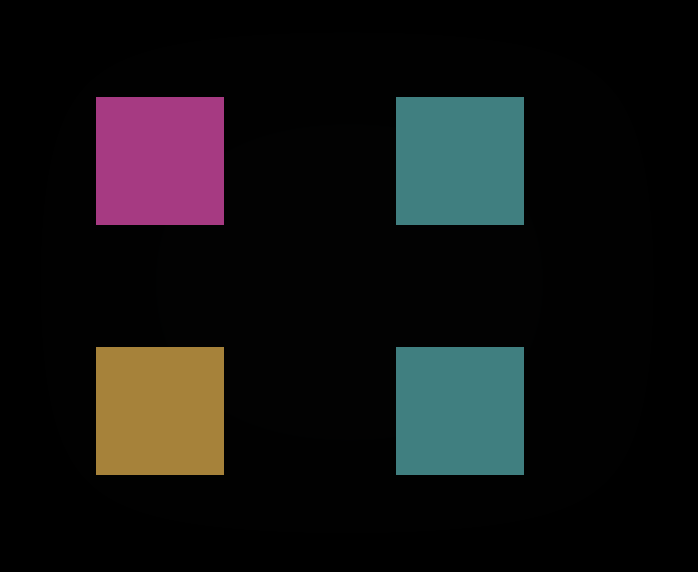}
  \label{fig:hm0_gt}}
  \\
     \subfloat[Multi-scale with TV-L1 energy]{
    \includegraphics[width=0.3\textwidth]{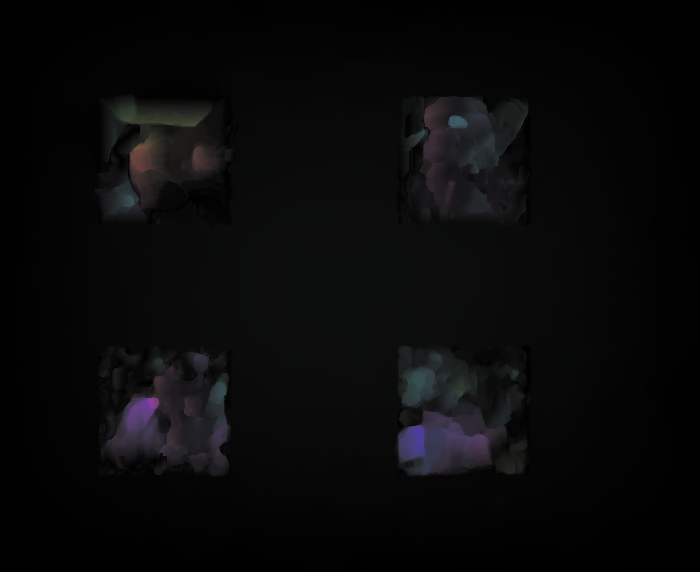}
  \label{fig:hm0_multi}}
  \subfloat[Ours with TV-L1 energy]{
    \includegraphics[width=0.3\textwidth]{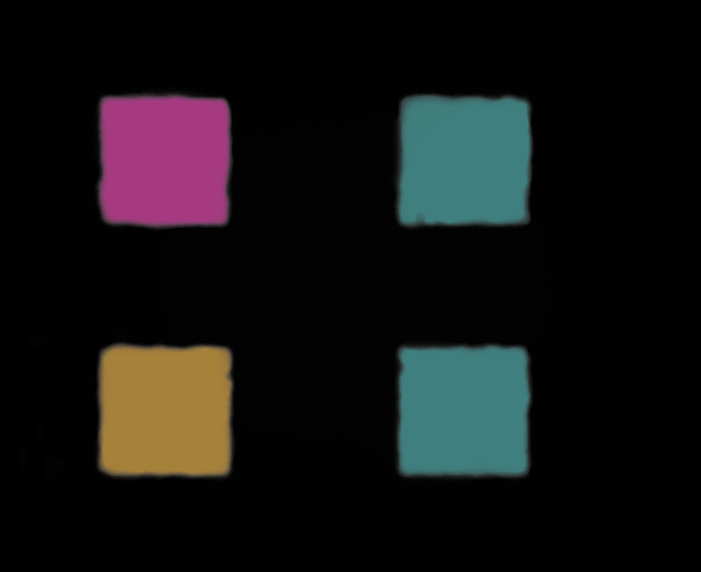}
  \label{fig:hm0_faldoi}}
  \\
  \subfloat[DeepFlow~\cite{Weinzaepfel2013}]{
    \includegraphics[width=0.3\textwidth]{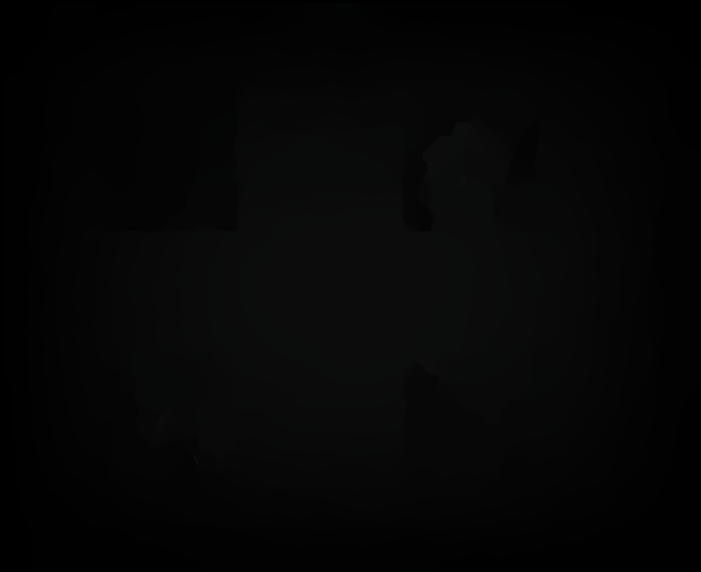}
  \label{fig:hm0_df2}}
  \subfloat[LDOF~\cite{Brox_Malik_LD}]{
    \includegraphics[width=0.3\textwidth]{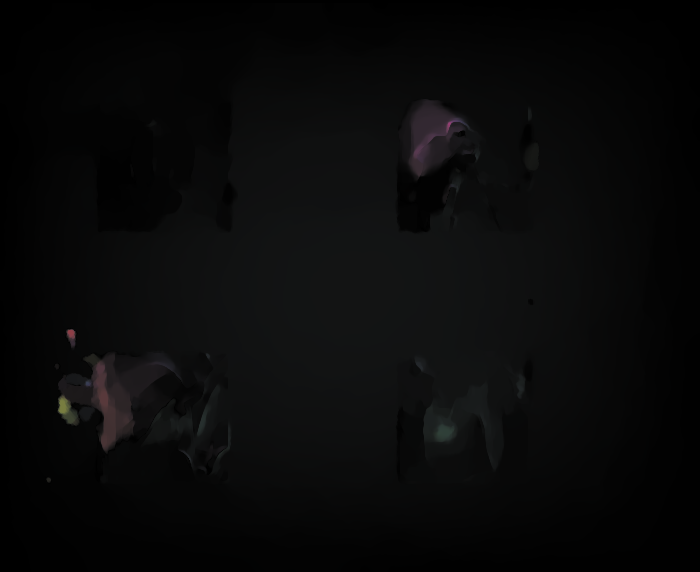}
  \label{fig:hm0_real_ldof}} 
  \caption{Large displacement on a composition made of the images \textit{Barbara}, \textit{Baboon}, \textit{Cameraman}, \textit{Lena} and \textit{Peppers}. It contains large diagonal translations and a slight deformation of the background. Five initial seeds have been manually selected and are shown in (a). The figures show 
  the differences among  Deepflow~\cite{Weinzaepfel2013} which is a multi-scale strategy using seeds similar to LDOF approach, the multi-scale approach for the TV-L1 energy, and  our proposal for minimizing the same energy. The last image shows the LDOF \cite{Brox_Malik_LD} result computed with its own seeds.
  }\label{fig:hypermacaco}
\end{figure*}

In the approach of Brox et al.~\cite{Brox_Malik_LD} or Weinzaepfel et al.~\cite{Weinzaepfel2013} the matches are precomputed and then added as a constraint to the energy term. 
Thanks to these matches the motion of small objects that disappear at the coarser scales is recovered. However, these approaches need 
a minimum density of sparse matches over the area of the small object in order to correctly capture large displacements, even if the matches are weighted strong enough and enough iterations are performed.
By contrast, our proposal only needs one single seed per area motion to recover the whole motion field. This is illustrated in Figure \ref{fig:hypermacaco} were our proposal is able to recover the four large displacements with just a single seed in each region while DeepFlow doesn't succeed. We also include the result obtained with LDOF~\cite{Brox_Malik_LD}, but in this case it has been obtained with their own seeds (using their originally binary code).

Recent optical flow datasets as MPI-Sintel \cite{Sintel} contain different and challenging effects, such as illumination changes, large displacements, blur, etc. In MPI-Sintel these effects have been artificially created to provide naturalistic video sequences. However, the evaluation on these datasets does not take into account the robustness to the shot noise that appears in any real sequence, being noise one of the main limitations to any imaging system. Thus, we evaluate the robustness of several approaches to noise. To this goal, we corrupt the clean images from~\cite{Sintel}  with additive white Gaussian noise of standard deviation $\sigma$, for several values of $\sigma$. Sparse-to-dense techniques are very dependent of the initial seeds that are used to obtain a dense optical flow. It is important to note that our proposal works even with an extremely sparse set of initial seeds. This fact allows us to choose the best matching method  according to the image peculiarities, without caring that much about the density of the correspondences. In particular, for highly noisy images SIFT correspondences are more robust than DeepMatching ones. Table \ref{table:noise} shows a comparative of our method with different optical flow estimation methods for different levels of Gaussian noise. We use the $TV$-$L_1$ functional with SIFT matches. 
As shown in the table, the multi-scale method  provides the best optical flow estimation in noisy images.  For noise levels of standard deviation greater or equal to 10 and 20 respectively, our method produces better results than EpicFlow and DeepFlow.

\begin{table*}
\begin{center}
\begin{tabular}{|lllll|}
\hline
$\sigma$ &  10 & 20 & 30 & 40 \\
\hline\hline
$DeepFlow$ without matches (pure multi-scale) & 0.7468  & 1.0353   & 1.3832  & 1.4808 \\
$DeepFlow$ \cite{Weinzaepfel2013}  & 0.7766  & 1.5796   & 2.6128  & 4.3918 \\
$EpicFlow$ \cite{EpicFlow} & 1.1418  & 1.6654   & 2.3386  & 3.2328 \\
$Ours$ ($TV_{\ell_2}$-$L1$) with SIFT matches  &  0.9486  & 1.5634  & 2.1090  & 2.9830\\
\hline
\end{tabular}
\end{center}
\caption{Results over a set of five samples from  the MPI-Sintel clean dataset with different standard deviation ($\sigma$) levels of Gaussian noise.
}
\label{table:noise}
\end{table*}

\section{Conclusions}
\label{sec:Conclusions}
 
We have proposed and provided an in-depth analysis of a method based on a variational formulation of the optical flow problem. 
The method works at the original scale of the image and finds a good local
minimum of any energy functional using an adaptive coordinate descent strategy
guided by a sparse set of initial matches.  This is a general technique that
consistently outperforms the multi-scale strategy for the same energy
functional.  
With respect to alternative techniques that also include sparse matches in any energy functional, our performance is comparable to DeepFlow \cite{Weinzaepfel2013} and superior to  LDOF \cite{Brox_Malik_LD} while being more robust to a low density of matches, high levels of noise and outliers in the matches.
The only requirement is that at least one correct match is given for each object in motion. 
For best overall results, we propose to use an energy with advanced data and regularization terms.  
Namely, we chose a smooth variant of the Census transform with a non-local $TV$ regularization, providing robustness to illumination changes and occlusions while handling motion discontinuities. 
We present accurate quantitative and qualitative results that are comparable with state-of-the-art methods. 
As future work we plan to model the occlusions in the functional to reduce the halo effect in occluded regions.

\appendix
\section{Minimizing the energy}\label{appendix:optalgos}

The numerical minimization algorithm for the general energy~(\ref{eq:energy}) is obtained in this paper by decoupling both terms.
We linearize the image $I_{t+1}$ near a given optical flow $\u_0=(u_{0,1},u_{0,2})$ and make the following approximation $I_{t+1}(\x+\u(x)) \approx I^{lin}_{t+1}(\x+\u(\x))$, where 
$$ I^{lin}_{t+1}(\x+\u(\x)) = I_{t+1}(\x+\u_0(\x)) + I^x_{t+1}(\x+\u_0(\x)) (u_1 - u_{0,1})(\x) + I^x_{t+1}(\x+\u_0(\x)) (u_2 - u_{0,2})(\x),$$
and $I^x_{t+1}$, $I^y_{t+1}$ denote the partial derivatives of $I_{t+1}$ with respect to $x$ and $y$ respectively. 
Let us recall that the two data terms $E_D$ that we have considered in Sect.~\ref{ssec:severalEnergies} depend on $I_t(\x)$ and $I_{t+1}(\x+\u(\x))$; we will denote as  $E_{D,lin}$ the same data term but depending on $I_t(\x)$ and  $ I^{lin}_{t+1}(\x+\u(\x))$. 
In order to decouple the fidelity term $E_{D,lin} (u)$ and the regularization term $E_{R} (\u)$  in~(\ref{eq:energy}), we introduce an auxiliary variable $\v$ representing the optical flow and we penalize its deviation from $\u$. Thus, the energy to minimize is
\begin{eqnarray}\label{eq:decoupled-energy}
   J(\u,\v) = J_{D,lin} (\v)  + \beta J_{R} (\u)  + \frac{1}{2\theta}\int_{\Omega}\|\u-\v\|^2,
\end{eqnarray}
depending on the two variables $\u, \v$, where $\theta>0$. The decoupled energy~(\ref{eq:decoupled-energy}) can be minimized by an alternating minimization procedure; alternatively fixing one variable and minimizing with respect to the other one. Sect. \ref{ssec:severalEnergies} presents the different possibilities for the energy.

\begin{enumerate}
\item For $\v$ fixed, let us consider each of the two different regularization terms, $J^1_R(\u)$ and $J^2_R(\u)$, presented in Sect. \ref{ssec:severalEnergies}.
\begin{itemize}
    
\item[1.1.] In the case of $J^1_R(\u)$, 
we reformulate the problem as a min-max problem incorporating the dual variables. Then, our minimization problem can be solved as a saddle-point problem. Following the notation of Osher \etal~\cite{gilboa2008nonlocal}, for $\v=(v_1,v_2)$ fixed, we solve
\begin{eqnarray} \label{eq:AP_EqR1_dual}
    \int_\Omega \int_\Omega  \omega(\x,\y) (u_i(\x) - u_i(\y)) p(\x,\y) d\y d\x  +  \frac{1}{2\theta}\int_{\Omega}\left(u_i -v_i\right)^{2} d\x,
\end{eqnarray}

for $i=1,2$, and $p$ is the dual variable defined on $\Omega \times \Omega$.
Let us explain it in detail. First, it is necessary to extend the notion of derivatives to a non-local framework. The non-local derivative can be written as 
\begin{eqnarray}
\partial_{y}{u_i(\x)} = \frac{u_i(\x) - u_i(\y)}{d(\x,\y)}
\end{eqnarray}
where $d(\x,\y)$ is a positive measure between two points $\x,\y$. By taking $d(\x,\y)$ such that $w(\x,\y) = d(\x,\y)^{-2}$, the non-local gradient $\nabla_{w}u_i(\x,\y)$ 
is defined as the vector of all partial derivatives:
\begin{eqnarray}
\nabla_{w}u_i(\x,\y) = (u_{i}(\x) - u_{i}(\y))\sqrt{w(\x,\y)} \quad  \x,\y \in \Omega.
\end{eqnarray}
Now, by writing $\vec{p} := p(\x,\y)$ for $(\x,\y)\in \Omega \times \Omega$, the non-local divergence $div_{w}\vec{p}(\x)$ 
is defined as the adjoint of the non-local gradient:
\begin{eqnarray}
\textnormal{div}_{w}\vec{p}(\x) = \int_{\Omega}  (p(\x,\y) - p(\y,\x))\sqrt{w(\x,\y)} d\y.
\end{eqnarray}

\begin{proposition}
\label{prop:nltv}
The solution of (\ref{eq:AP_EqR1_dual}) is given by the following iterative scheme
\begin{eqnarray}
p(\x,\y)^{n+1} & = & \frac{p(\x,\y)^{n} + \tau (\overline{u}_{i}^{n}(\x) - \overline{u}_{i}^{n}(\y)\sqrt{w(\x,\y)})}{1 + \tau |\nabla_{w}u_{i}(\x,\y)|}\\
u_{i}^{n+1}(\x) & = & u_{i}^{n}(\x) - \sigma \left(\frac{(u_{i}^{n}(\x) - v_{i}(\x))}{\theta} - \textnormal{div}_{w} \vec{p}(\x)\right)\\
\overline{u}_{i}^{n+1}(\x) & = & 2u_{i}^{n+1}(\x) -u_{i}^{n}(\x)
\end{eqnarray}
where $u_{i}$ is the primal variable and $\vec{p}$ is the dual variable. 
\end{proposition}

        \item[1.2.] In the case of $J^2_R(\u)$ we use the primal-dual algorithm that Chambolle proposed to minimize the ROF model \cite{Chambolle2010} and which is based on a dual formulation of the $TV$. Then, our minimization problem can be solved as a saddle-point problem. For $\v$ fixed, we solve
  \begin{eqnarray}\label{tvl2_Rof_Model}
   \min_\u \max_\mathbf{\xi} \int_{\Omega}  \langle D \u,\xi\rangle d\x + \int_{\Omega} \frac{1}{2\theta}\|\u -\v\|)^{2} d\x
\end{eqnarray}
where the dual variables are
$
\mathbf{\xi} =   \left( \begin{array}{cc}
\xi_{11} & \xi_{12}  \\
\xi_{21} & \xi_{22} \end{array} \right)
$ and satisfy  $||\xi||_{F} \leq 1$.

\begin{proposition}
\label{prop:tvcoupled}
\label{min-u}
The solution of (\ref{tvl2_Rof_Model}) is given by the following iterative scheme
\begin{eqnarray}
\xi_{i1}^{n+1} & = & \frac{\xi_{i1}^{n} + \tau \overline{u}_{ix}^{n}}{\max(1,||\xi ||_{2})}, \qquad
\xi_{i2}^{n+1}  =  \frac{\xi_{i2}^{n} + \tau \overline{u}_{iy}^{n}}{\max(1,||\xi ||_{2})},\\
u_{i}^{n+1} & = & u_{i}^{n} - \sigma \left(\frac{(u_{i}^{n} - v_{i})}{\theta} - \textnormal{div}\left(\xi_{i1}^{n}, \xi_{i2}^{n}\right)\right),\\
\overline{u}_{i}^{n+1} & = & 2u_{i}^{n+1} -u_{i}^{n}, 
\end{eqnarray}
 where $i=1,2$.
\end{proposition}
\end{itemize}

\item For $\u$ fixed, let us consider each of the two different data terms, $J^1_D(\v)$ and $J^2_D(\v)$, presented in Sect. \ref{ssec:severalEnergies}. 
\begin{itemize}
    \item[2.1.] Case $J^1_D(\v)$, Li and Osher \cite{li2009} present a simple algorithm to find the optimal value of the function $E(x) = \sum\limits_{i}^{n} w_i |x-a_i| + F(x)$ when the $w_i$ are non-negative and $F$ is strictly convex. If $F$ is also differentiable and $F'$ is bijective, it is possible to obtain an explicit formula in terms of the median. For $\u$ fixed, we solve
    \begin{eqnarray} \label{eq:AP_ED}
  \int_\Omega C(\v,\x) d\x + \frac{1}{2\theta} \int_{\Omega} \|\u -\v\|^{2} d\x.
\end{eqnarray}
    Following the ideas of \cite{vogel2013evaluation}, we solve the discrete version of this problem. Due to the isotropy of the quadratic term, the optimal solution of $C(\v,\x)$ can be obtained solving a one dimensional problem. In particular, setting $\v = \hat{\v} + \delta \frac{\nabla I(x + v_o)}{|\nabla I(x + v_o)|} + \delta \frac{\nabla^{+} I(x + v_o)}{|\nabla^{+} I(x + v_o)|}$ being $\nabla^{+}I$ an orthogonal vector to the gradient, where $\delta$ is our new variable. Then, we minimize over $\delta$
    \begin{eqnarray}\label{eq:Ap_csad_model}
     \frac{1}{2\tau}\delta^2 +  \lambda \int_\Omega |\nabla I_{t+1}(\x + \hat{\v}_o)|\left|G(\hat{\v}) + \delta\right|  d\y 
    \end{eqnarray}
    where $$
    G_{\y}(\hat{\v}) = \frac{I_t(\x) - I_t(\y) - I_{t+1}(\x + \hat{\v}_o) + I_{t+1}(\y + \hat{\v}_o) + (\hat{\v}-\hat{\v}_o)^{T}\nabla I_{t+1}(\x + \hat{\v}_o)}{|\nabla I_{t+1}(\x + \hat{\v}_o)|}
    $$
    \begin{proposition}\label{prop:min-data-csad}
    The minimum of (\ref{eq:Ap_csad_model}) with respect to $\delta$ is
    \begin{eqnarray}
            \delta^{*} = \textnormal{median} \{ b_1,...,b_n,a_0,...a_n \}
    \end{eqnarray}
where $b_i = -G_{i}(\hat{\v})$ and $a_i = (n-2i)\lambda|\nabla I_{t+1}(\x + \v_o)|$ for all the discrete neighbors $i$ (corresponding to $\y$ above), where $n$ is the number of points in the discrete neighborhood. 
    \end{proposition}

    \item[2.2.] Case $J^2_D(\v)$. Notice that this term is a particular case of the previous data term. The  
    functional to minimize 
    \begin{eqnarray}
   \int_{W}  \lambda |\rho(\v)|  +\frac{1}{2\theta}\int_{W} \|\u -\v\|^{2} d\x,\label{Data_SubProblem_Model}
\end{eqnarray}
where $\rho(\v) = I_{t+1}(\x,\v_o) + \langle \nabla I_{t+1}(\x + \v_o),(\v - \v_o) \rangle  -  I_{t}(\x)$, does not depend on spatial derivatives on $\v$. Then, a simple thresholding step gives an explicit solution \cite{Zach}. 

\begin{proposition}
\label{prop:min-data-bca}
The minimum of (\ref{Data_SubProblem_Model}) with respect to $\v$ is 
\begin{eqnarray}              
\label{v1_eq}
\v = \u + 
\left\{
\begin{array}{lll}
\lambda \theta  \nabla I_{t+1} \quad \quad \quad  \; \;\; \mathrm{  if} \quad  \rho(\u) < -\lambda\theta|\nabla I_{t+1}|^{2}\\
- \lambda \theta  \nabla I_{t+1} \quad \quad \quad \;  \mathrm{if }  \quad \rho(\u) > \; \lambda\theta|\nabla I_{t+1}|^{2}\\
-\rho(\u)\frac{\nabla I_{t+1}}{|\nabla I_{t+1}|^{2}} \quad \quad  \mathrm{  if} \quad  |\rho(\u)| \leq \lambda\theta|\nabla I_{t+1}|^{2}
\end{array}
\right.
\end{eqnarray}
\end{proposition}
\end{itemize}

\end{enumerate}
\begin{algorithm}
 \SetAlgoLined
 \SetKwInOut{Input}{Input}\SetKwInOut{Output}{Output}
 \Input{Two consecutive frames $I_{1},I_{2}$, and  an initial optical flow $\u_{0}$} 
  \Output{Flow field $\u$}
 \BlankLine
  Initialize $\xi=\v=\vec{0}$\;
  Initialize $\u = \u_{0}$\;
  \For{$w \leftarrow$ 1 to $N_{warps}$}
  {
    Compute $I_{t+1}(\x+ \v_{0}(\x))$, 
    $I^{x}_{t+1}(\x+ \v_{0}(\x))$,  
    $I^{y}_{t+1}(\x+ \v_{0}(\x))$, 
   using bicubic interpolation\;
    \While{$n<N_\mathrm{max}$ or $tol<error$}{

      Compute $\v$ via Prop.~\ref{prop:min-data-csad} or Prop.~\ref{prop:min-data-bca} \tcp*[r]{for $CSAD$ or $L1$}
      Compute $\xi$ or $p$ and $\u$ via Prop.~\ref{prop:nltv} or Prop.~\ref{prop:tvcoupled} \tcp*[r]{for $NLTV$  or $TV_{\ell_2}$-$L1$}
     }
 } 
 \caption{Global Minimization to compute the final optical flow.}
 \label{algorithm:global_min} 
\end{algorithm}

\section{Implementation details}\label{appendix:impldetails}
Our code is written in C. The numerical scheme to solve the functional $E(u)$, in both steps, is based on the implementation of \cite{Palomares2014}. Image warpings use bicubic interpolation. The image gradient is computed using centered-derivatives. Input images have been normalized between $[0,1]$. The algorithm parameters are initialized with the same default setting for all the experiments. Both time steps are set to $\tau = \sigma = 0.125$ to ensure  convergence. As stopping criterion, the optical flow method uses the infinite-norm between two consecutive values of $u$ with a threshold of $0.01$. The coupling parameter $\theta$ is set to $0.3$. The smoothness term weight $\beta$ is set to $1/40$ for the $TV_{\ell_2}$-$L1$ functional and $\beta = \frac{N-1}{80}$ for the $NLTV$-$CSAD$ one,  as suggested by \cite{vogel2013evaluation}, where $N$ is the cardinality of the neighborhood considered in the $CSAD$ term (we use a neighborhood of $7\times7$ in the data term and then $N=49$) and we fixed $\sigma_c=2$ and $\sigma_s=2$ for the spatial an color domain of the $NLTV$ term. For the iterated faldoi strategy we set $MAX\_IT$ to 3.
The size of the patch $\cal{P}$ in the local minimization is $11\times 11$. The complexity of our algorithm is $\mathcal{O}(n)$, where $n$ is the number of pixels of the image frame. The basic faldoi algorithm takes around 20 seconds for $TV_{\ell_2}$-$L1$ energy and around 10 minutes for $NLTV$-$CSAD$  over an $MPI-Sintel$ image. Notice that our algorithm using $NLTV$-$CSAD$ energy is very slow, especially because our implementation does not use  parallized code. As $NLTV$-$CSAD$ can be easily parallelized, it should take the same time for both functionals using a GPU implementation.
\begin{acknowledgements}
The authors would like to thank Gabriele Facciolo for all the discussions with him about the subject of this paper
\end{acknowledgements}

\bibliographystyle{spmpsci}      
\bibliography{ref}
\end{document}